\documentclass[nohyperref]{article}

\usepackage{subcaption}
\usepackage{multirow}
\usepackage{microtype}
\usepackage{graphicx}
\usepackage{booktabs} 
\usepackage{diagbox}
\usepackage{lipsum}  

\newcommand{\I}{\boldsymbol{I}}

\newcommand{\Z}{\boldsymbol{Z}}

\newcommand{\x}{\boldsymbol{x}}

\newcommand{\z}{\boldsymbol{z}}

\newcommand{\Zb}{{\mathbf Z}}

\newcommand{\vb}{{\boldsymbol v}}

\newcommand{\phib}{{\boldsymbol{\phi}}}
\newcommand{\psib}{{\boldsymbol{\psi}}}

\newcommand{\thetab}{{\boldsymbol {\theta}}}

\newcommand{\beq}{\begin{equation}}
\newcommand{\eeq}{\end{equation}}
\newcommand{\beqa}{\begin{eqnarray}}
\newcommand{\eeqa}{\end{eqnarray}}

\newcommand{\heun}{{Heun's $2^{\text{nd}}$ order method}}
\newcommand{\code}{https://github.com/sangyun884/fast-ode}
\usepackage{hyperref}
\usepackage{enumitem}

\usepackage[accepted]{icml2023}

\usepackage{amsmath}
\usepackage{amssymb}
\usepackage{mathtools}
\usepackage{amsthm}

\usepackage[capitalize,noabbrev]{cleveref}

\theoremstyle{plain}
\newtheorem{theorem}{Theorem}
\newtheorem{proposition}[theorem]{Proposition}

\theoremstyle{definition}

\theoremstyle{remark}

\usepackage[textsize=tiny]{todonotes}

\icmltitlerunning{Minimizing Trajectory Curvature of ODE-based Generative Models}

\begin{document}

\twocolumn[
\icmltitle{Minimizing Trajectory Curvature of ODE-based Generative Models}




\icmlsetsymbol{equal}{*}

\begin{icmlauthorlist}
\icmlauthor{Sangyun Lee}{soongsil}
\icmlauthor{Beomsu Kim}{kaist}
\icmlauthor{Jong Chul Ye}{kaist}


\end{icmlauthorlist}

\icmlaffiliation{soongsil}{Soongsil University}
\icmlaffiliation{kaist}{KAIST}

\icmlcorrespondingauthor{Jong Chul Ye}{jong.ye@kaist.ac.kr}

\icmlkeywords{Machine Learning, ICML}

\vskip 0.3in
]



\printAffiliationsAndNotice{}  

\begin{abstract}
Recent ODE/SDE-based generative models, such as diffusion models, rectified flows, and flow matching, define a generative process as a time reversal of a fixed forward process. Even though these models show impressive performance on large-scale datasets, numerical simulation requires multiple evaluations of a neural network, leading to a slow sampling speed. We attribute the reason to the high curvature of the learned generative trajectories, as it is directly related to the truncation error of a numerical solver. Based on the relationship between the forward process and the curvature, here we present an efficient method of training the forward process to minimize the curvature of generative trajectories without any ODE/SDE simulation. Experiments show that our method achieves a lower curvature than previous models and, therefore, decreased sampling costs while maintaining competitive performance. Code is available at \href{\code}{\code}.
\end{abstract}

    \vspace*{-0.5cm}
\section{Introduction}
    \vspace*{-0.1cm}
Many machine learning problems can be formulated as discovering the underlying  distribution from observations. Owing to the development of deep neural networks, deep generative models exhibit superb modeling capabilities.

Classically, Variational Autoencoders (VAE)~\citep{kingma2013auto}, Generative Adversarial Networks (GAN)~\citep{goodfellow2014generative}, and invertible flows~\citep{rezende2015variational} have been extensively studied.
However, each model has its drawback.
GANs have dominated image synthesis for several years~\citep{karras2019style,brock2018large,karras2020analyzing}, but carefully selected regularization techniques and hyperparameters are needed to stabilize training~\citep{miyato2018spectral,brock2018large},
and their performance often does not transfer well to other datasets.
Invertible flows enable exact maximum likelihood training, but the invertibility constraint significantly restricts the architecture choice, which means that they cannot benefit from the development of scalable architectures.
VAEs do not suffer from the invertibility constraint, but their sample quality is not as good as other models. 

Apart from the vanilla VAEs, recent studies utilize their hierarchical extensions~\citep{child2020very,vahdat2020nvae} as they offer more expressivity to both inference and generative components by assuming nonlinear dependencies between latent variables. However, they often have to rely on heuristics such as KL-annealing or gradient skipping due to training instabilities~\citep{child2020very,vahdat2020nvae}.
Although continuous normalizing flows~\citep{chen2018neural} do not suffer from the invertibility constraint and can be trained on a stationary objective function, training requires simulating ODEs, which prevents them from being applied to large-scale datasets.


Recent ODE/SDE-based approaches attempt to settle these issues by defining the generative process as a time reversal of a fixed forward process. Diffusion models~\citep{song2019generative,song2020score,ho2020denoising,sohl2015deep} define the generative process as a time reversal of a forward diffusion process, where data is gradually transformed into noise. By doing so, they can be trained on a stationary loss function~\citep{vincent2011connection} without ODE/SDE simulation. Moreover, they are not restricted by the invertibility constraint and can generate high-fidelity samples with great diversity, allowing them to be successfully applied to various datasets of unprecedented scales~\citep{saharia2022photorealistic,ramesh2022hierarchical}. 
Rectified flow~\citep{liu2022flow} provides a different perspective on this model class. From this viewpoint, the training of diffusion models can be seen as matching the forward and reverse vector fields. Since stochasticity is not a root of the success of these models~\citep{karras2022elucidating} and rectified flow offers an alternative perspective that is fully explained under the ODE scheme, we hereafter refer to these types of models as \textit{ODE-based generative models}.
If necessary, a generative ODE can be easily converted to an SDE and vice versa \citep{song2020score}.

\begin{figure*}[t]
    \centering
    \includegraphics[width=0.73\linewidth]{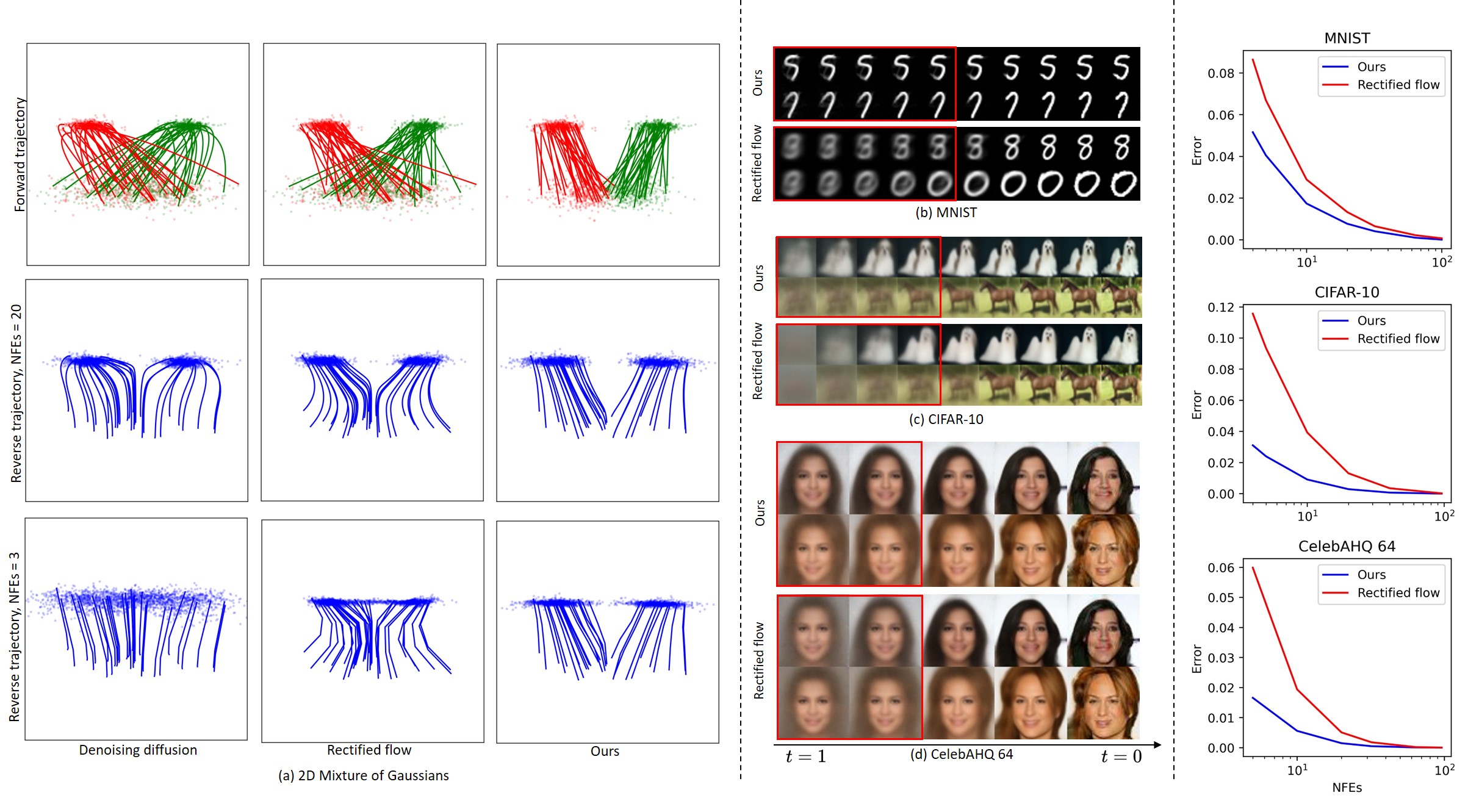} 
    \vspace*{-0.5cm}
    \caption{Forward and reverse trajectories of denoising diffusion model~\citep{ho2020denoising}, rectified flow~\citep{liu2022flow}, and our method on 2D dataset (\textit{left}). The intersection between forward trajectories makes reverse trajectories 
 collapse toward the average direction, resulting in increased curvature and suboptimal sample qualities with a limited number of function evaluations (NFE). In contrast, our approach successfully \textit{unties} the crossover between forward trajectories, leading to low-curvature reverse trajectories. This phenomenon also holds true in high-dimensional spaces, as demonstrated by reverse process visualization on MNIST, CIFAR-10, and CelebAHQ (64 $\times$ 64) datasets (\textit {middle}). As a result, our method makes less truncation error when the number of function evaluations (NFE) is small (\textit {right}).}
     \vspace*{-0.5cm}
    \label{fig:main}
\end{figure*}

However, drawing samples from ODE-based generative models requires multiple evaluations of a neural network for accurate numerical simulation, leading to slow sampling speed. While many studies have attempted to develop fast samplers for pre-trained models~\citep{lu2022dpm,zhang2022fast}, there seems to be a limit to lowering the costs.
We attribute the reason to the high curvature of the learned generative trajectories. The curvature is intriguing since it is directly related to the truncation error of a numerical solver. Intuitively, zero curvature means that generative ODEs can be accurately solved with only one function evaluation.
Since a generative process is a time reversal of the forward process, it is evident that its curvature is also somehow determined by the forward process, but the exact mechanism is yet unexplored. We find that the rectified flow perspective offers an interesting insight into the relationship between the forward process and the curvature. Based on our observation, we propose an efficient method of training the forward process to reduce curvature. Specifically, our contributions are as follows: 

\begin{itemize}
    \item {We investigate the relationship between the forward process and curvature from a rectified flow perspective. We find that the degree of intersection between forward trajectories is positively related to the curvature of generative processes.}
    \item {We propose an efficient method of learning the forward process to reduce the degree of intersection between forward trajectories without any ODE/SDE simulation. We show that our method can be seen as a $\beta$-VAE~\citep{higgins2016beta} with a time-conditional decoder.}
    \item {Experiments show that our method achieves lower curvature than previous models and, therefore, demonstrates decreased sampling costs while maintaining competitive performance.}
\end{itemize}

    \vspace*{-0.5cm}
\section{Background}
ODE/SDE-based generative models effectively model complex distributions by repeatedly composing a neural network, making trade-offs between execution time and sample quality.
In this paper, we focus on ODE-based generative models since they yield the same marginal distribution as SDEs while being conceptually simpler and faster to sample~\citep{song2020score}.

Different from continuous normalizing flows (CNF)~\citep{chen2018neural}, recent ODE-based models do not require ODE simulations during training and therefore are more scalable.
At a high level, they define a forward coupling $q(\x,\z)$ between data distribution $p(\x)$ and prior distribution $p(\z)$ and subsequently an interpolation $\x_t(\x,\z)$ for $t \in [0,1]$ between a pair $(\x,\z) \sim q(\x,\z)$ such that $\x_0(\x,\z) = \x$ and $\x_1(\x,\z) = \z$.
Training objectives are variants of the denoising autoencoder objective
\begin{equation}
    \min_\thetab \mathbb E_{t \sim U(0, 1)}\mathbb E_{\x, \z \sim q(\x, \z)}[\lambda(t)||\x - \x_\theta(\x_t(\x, \z),t)||_2^2],
    \label{eq:dsm}
\end{equation}
where $\lambda(t)$ is a weighting function. Here, a neural network $\x_\theta(\x_t,t)$ is trained to reconstruct the data $\x$ from the corrupted observation $\x_t$.
In the following, we briefly review two popular instances of such models: the denoising diffusion model and rectified flow. We refer the readers to Appendix~\ref{appendix:preliminaries} for a detailed background.

\paragraph{Denoising diffusion models}
Denoising diffusion models~\citep{ho2020denoising} employ the prior $p(\z) = \mathcal N(\mathbf 0, \I)$, forward coupling $q(\x,\z) = p(\x)p(\z)$, and a nonlinear interpolation
\begin{equation}
    \x_t(\x,\z) = \alpha(t)\x + \sqrt{1 - \alpha(t)^2}\z
    \label{eq:diffusion-interpolation}
\end{equation}
with a predefined nonlinear function $\alpha(t)$. $\lambda(t)$ is often adjusted to improve the perceptual quality for image synthesis~\citep{ho2020denoising}. Sampling can be done by solving probability flow ODEs~\citep{song2020score}.

\paragraph{Rectified flows}
However, the choice of Eq.~\eqref{eq:diffusion-interpolation} seems unnatural from a rectified flow perspective as it unnecessarily increases the curvature of generative trajectories.
In rectified flow~\citep{liu2022flow,liu2022rectified}, the intermediate sample $\x_t$ is rather defined as a linear interpolation
\begin{equation}
    \x_t(\x, \z) = (1-t)\x + t\z
    \label{eq:linear-interpolation}
\end{equation} 
since it has a constant velocity across $t$ for given $\x$ and $\z$. After training, sampling is done by solving the following ODE backward:
\begin{equation}
    d\z_t = \frac{\z_t - \x_\theta(\z_t,t)}{t}dt,
    \label{eq:flow-sampling}
\end{equation}
where $dt$ is an infinitesimal timestep. In the optima, Eq.~\eqref{eq:flow-sampling} maps $\z_1$ from $p(\z)$ to $\z_0$ following $p(\x)$.
Instead of predicting $\x$, \citet{liu2022flow} directly learns the velocity $\vb_\theta(\z_t, t) = \frac{\z_t - \x_\theta(\z_t,t)}{t}$.
The effectiveness of this sampler in reducing the sampling costs has been previously investigated in \citet{karras2022elucidating} under the variance-exploding context. Also, Eqs.~\eqref{eq:linear-interpolation} and \eqref{eq:flow-sampling} are a special case of flow matching~\cite{lipman2022flow}. See Appendix~\ref{appendix:flow-matching}. 
We build our method based on this framework, using the linear interpolation and the ODE in Eq.~\eqref{eq:flow-sampling} for sampling.

\section{Curvature Minimization}
\subsection{Curvature}
For a generative process $\Z = \{\z_t(\z)\}$ with the initial value $\z_1(\z) = \z$ , we informally define curvature as the extent to which the trajectory deviates from a straight path:
\begin{equation}
    C(\Z) = \mathbb E_t \left\|\z_1(\z) - \z_0(\z) - \frac{\partial}{\partial t}\z_t(\z) \right\|_2^2,
\end{equation}
which is equal to the straightness used in \citet{liu2022flow}.
The average curvature $\mathbb E_{\z \sim p(\z)}[C(\Z)]$ should be the main concern in designing the ODE-based models since it is directly related to the truncation error of numerical solvers. Zero curvature means the path is completely straight. Therefore, a {\em single} step of the Euler solver is sufficient to obtain an accurate solution.

Since a generative process is a time reversal of the forward process, its curvature is determined by the forward process. As an illustrative example, consider a generative ODE that is trained on Eq.~\eqref{eq:dsm}.
In the optima, $\x_{\thetab}(\z_t, t)$ is a minimum mean squared error estimator $ \mathbb E[\x | \x_t = \z_t]$, and the average curvature of the generative processes governed by Eq.\eqref{eq:flow-sampling} becomes
\begin{equation}
    \mathbb E_{\z, t}\left\|\z_1(\z) - \z_0(\z) - \frac{1}{t} \z_t(\z) + \frac{1}{t} \mathbb E[\x | \x_t = \z_t(\z)] \right\|_2^2,
    \label{eq:optim-curvature}
\end{equation}
which is a function of the posterior $q(\x | \x_t)$. Since we define $\x_t$ as an interpolation between $\x$ and $\z$, the posterior is determined by the forward coupling $q(\x, \z)$.
In previous work, $q(\x, \z)$ is fixed, and so is the curvature of the generative process in optima.
In the following, we further examine the relationship between forward coupling and curvature and show that we can improve the curvature by finding better $q(\x, \z)$.

\subsection{Curvature and the degree of intersection}

Specifically, we observe that Eq.~\eqref{eq:optim-curvature} is related to the degree of intersection of the forward trajectories
\begin{equation}
    I(q) = \mathbb E_{t, \x, \z \sim q(\x, \z)}  [|| \z - \x - \mathbb E[\z - \x | \x_t(\x, \z)] ||_2^2],
    \label{eq:intersection}
\end{equation}
which becomes zero when there is no intersection at any $\x_t$. As shown in Fig.~\ref{fig:main}, the intersection between forward trajectories makes the reverse vector field collapse toward the average direction, leading to high curvature. As the degree of intersection decreases, the reverse paths are gradually straightened. When $I(q) = 0$, the posterior $q(\x | \x_t)$ becomes a Dirac delta function, $\mathbb E[\x | \x_t = \z_t(\z)] = \z_0(\z)$ for every $t$, and Eq.~\eqref{eq:optim-curvature} becomes zero, i.e., the paths are completely straight.
Therefore, it is natural to seek a forward coupling $q(\x, \z)$ that minimizes Eq.~\eqref{eq:intersection}.
We can estimate Eq.~\eqref{eq:intersection} by minimizing the following upper bound with respect to $\thetab$.

\begin{figure}[!hbt]
    \centering
    \includegraphics[width=0.75\linewidth]{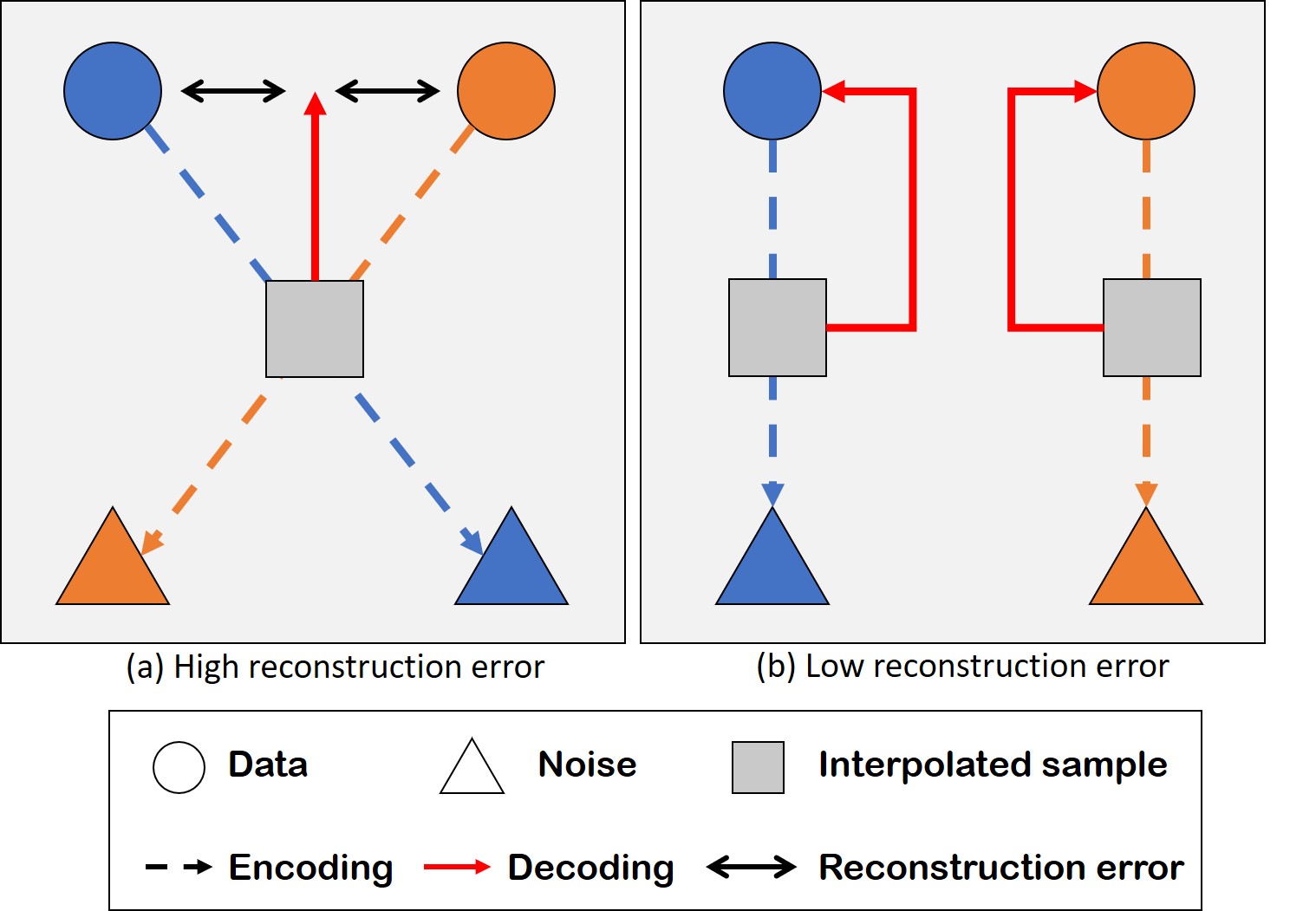} 
    \caption{Reconstruction error is (a) high when forward trajectories intersect (b) and low when they do not.}
    \label{fig:intuition}
\end{figure}

\begin{proposition}
Let $\x_t(\x, \z)$ be the linear interpolation defined as Eq.~\eqref{eq:linear-interpolation}. Then, we have

\begin{equation}
I(q) \leq \mathbb E_{t, \x, \z \sim q(\x, \z)}\left[ \frac{1}{t^2} ||\x-\x_\thetab(\x_t, t)||_2^2\right].
\label{eq:loss-theta}
\end{equation}

The bound is tight when $\x_\thetab(\x_t,t) = \mathbb E[\x | \x_t]$.
\begin{proof}[Sketch of Proof]
Using Eq.~\eqref{eq:linear-interpolation}, we obtain $\z - \x = (\x_t - \x) / t$. Plugging it into Eq.~\eqref{eq:intersection}, we have
\begin{equation}
    I(q) = \mathbb E_{t, \x, \z \sim q(\x, \z)}  [|| \frac{1}{t}(\x - \mathbb E[\x | \x_t]) ||_2^2],
\end{equation}
which is bounded by Eq.~\eqref{eq:loss-theta}.  
\end{proof}
\end{proposition}
For the independent coupling $q(\x, \z) = p(\x) p(\z)$, the upper bound of $I(q)$ coincides with Eq.~\eqref{eq:dsm} with $\lambda(t) = 1/t^2$, which is the training loss of \citet{liu2022flow}. In this sense, \citet{liu2022flow} estimates the upper bound of the degree of intersection of the independent coupling but does not really minimize it. Intuitively, the degree of intersection can be measured by the reconstruction error of an optimal decoder since the decoding is more difficult when multiple inputs are encoded into a single point. See Fig.~\ref{fig:intuition} for an illustration.

\subsection{Parameterizing $q(\x, \z)$}
After estimating $I(q)$ by updating $\thetab$, we search for $q$ that minimizes $I(q)$.
Although there are many ways to solve this optimization problem, there are two practical considerations. First, the optimization needs to be efficient.
Moreover, $q(\z|\x)$ should define a smooth map from $\x$ to $\z$ since we have to approximate $\mathbb E[\x | \x_t]$ using a neural network with finite capacity in practice.
Therefore, we propose to parameterize the coupling as a neural network $q_\phib(\x,\z) = q_\phib(\z | \x)p(\x)$, where we define $q_\phib(\z | \x)$ as a Gaussian distribution. With $q_\phib(\z) = \int q_\phib(\x, \z)d\x$ and a weight $\beta$, we optimize
\begin{align}
    \min_\phib I(q_\phib) + \beta D_{KL}(q_\phib(\z) || p(\z)).
    \label{eq:loss-q}
\end{align}
The second KL term ensures $q_\phib (\x, \z)$ is a valid coupling between $p(\x)$ and $p(\z)$. See Appendix~\ref{appendix:loss} for more details.

\begin{figure}[!hbt]
    \centering
    \includegraphics[width=0.7\linewidth]{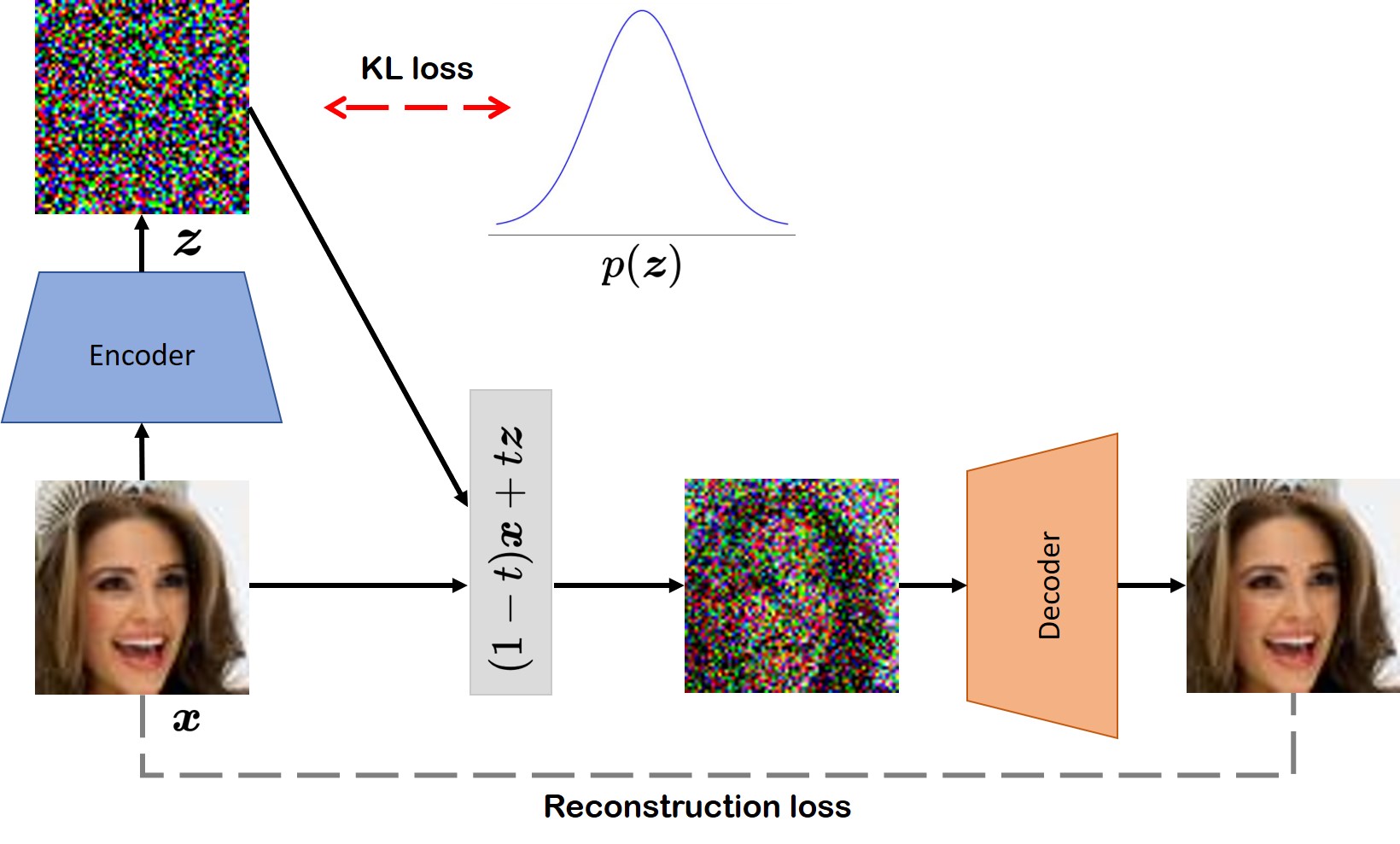} 
    \caption{A visual schematic of the proposed method.}
    \label{fig:method}
\end{figure}

\paragraph{Joint training}
In practice, we jointly minimize Eqs.~\eqref{eq:loss-theta} and \eqref{eq:loss-q} with respect to both $\theta$ and $\phi$. This leads to our loss function
\begin{align}
\nonumber
    \min_{\thetab, \phib}\ \mathbb E_{t, \x, \z \sim q_\phib(\x, \z)}[ \frac{1}{t^2} ||\x-\x_\thetab(\x_t(\x, \z), t)||_2^2
    \\ + \beta D_{KL}(q_\phib(\z | \x) || p(\z))],
    \label{eq:joint-loss}
\end{align}
which resembles the $\beta$-VAE objective~\citep{higgins2016beta} in that Eq.~\eqref{eq:joint-loss} reduces to the $\beta$-VAE loss if we fix $t$ to $1$. Since the decoder $\x_\theta$ is conditioned on time step, ODE-based models can synthesize higher quality samples than $\beta$-VAEs by iteratively refining the blurry initial predictions.  From this viewpoint, previous methods \cite{liu2022flow,ho2020denoising} can be seen as degenerate cases where the encoder $q_\phib(\z|\x)$ collapses into the prior by setting $\beta \rightarrow \infty$.
See Fig.~\ref{fig:method} for a visual schematic of our method.

\section{Related Works}
\paragraph{Alternative forward processes}
There have been several approaches to finding alternative forward processes for diffusion models. It has been demonstrated that other types of degradation, such as blurring, masking, or pre-trained neural encoding, can be used for the forward process~\citep{rissanen2022generative,lee2022progressive,hoogeboom2022blurring,daras2022soft,gu2022f,bansal2022cold}. However, they are either purely heuristic or rely on an inductive bias that is not necessarily well-supported by theory.

\paragraph{Learning forward process}
A few studies attempted to learn the forward process. \citet{kingma2021variational} proposed to learn the signal-to-ratio function of the forward process jointly with generative components. However, the inference model of \citet{kingma2021variational} is linear and thus has limited expressivity. \citet{zhang2021diffusion} proposed nonlinear diffusion models, where the drift function of the forward SDEs are neural networks. Although they introduce more flexibility in inference models, training requires the simulation of forward/reverse SDEs, which causes a significant computational overhead. Our method possesses the advantages of both methods. Our inference model is expressive since we set $q(\z | \x)$ as a neural network. Since we define $\x_t$ as an interpolation between $\x$ and $\z$ and $q_\phi(\x_t | \x)$ as a Gaussian distribution, sampling is done with one forward pass for an arbitrary $t$, enabling efficient training as in previous methods~\citep{ho2020denoising,liu2022flow}. Moreover, even though the nonlinear forward process appeared to improve the sampling efficiency of diffusion models~\cite{zhang2021diffusion}, the exact mechanism of the improved sampling speed was vague. In this paper, we convey a clear motivation for learning the forward process by revealing the relationship between the forward process and the curvature of the generative trajectories.
\paragraph{Fast samplers}
Accelerating the sampling speed of diffusion models is an active research topic, which is often tackled by developing fast solvers~\citep{lu2022dpm,zhang2022fast}. Our work is in an orthogonal direction since they focus on taming the high curvature ODEs while we aim to minimize the curvature itself. We expect the effect of these methods to be additive to ours and leave the detailed investigation for future work.
\paragraph{Straightness of Neural ODEs}
The importance of the straightness of neural ODEs in reducing the sampling cost has been previously discussed. Based on Benamou-Brenier formulation of the optimal transport problem~\citep{benamou2000computational}, \citet{finlay2020train} regularized the norm of the vector field of the CNFs to encourage the straightness, which is later generalized in \citet{kelly2020learning} where the norm of the $K$-th order derivative is minimized. Since CNFs are trained on the maximum likelihood objective, any vector field that defines the transport map from $p(\z)$ to $p(\x)$ is optimal, and it is therefore possible to narrow down the search space by utilizing the additional constraint without drastically compromising the performance. In contrast, recent ODE-based generative models~\citep{song2019generative,ho2020denoising,liu2022flow} train the neural ODE to match a pre-defined forward flow using Eq.~\eqref{eq:dsm}. Thus the solution is unique, and any additional regularization makes models deviate from optima.

\paragraph{Optimizing coupling}
Concurrent with our work, \citet{pooladian2023multisample} proposed to optimize $q(\x, \z)$ and showed several desirable properties such as improved sampling efficiency and reduced gradient variance during training.
While we parameterize $q(\x, \z)$ as a neural network, they construct a doubly-stochastic matrix for $q(\x, \z)$ and apply computational methods to find the optimal coupling between two empirical distributions.
In practice, the optimization is done either with heuristics or using mini-batch samples every iteration due to the computational burden.
They showed that in an ideal case with infinite batch size, 1) $I(q)$ goes to zero, 2) $C(\Zb)$ becomes zero, and 3) the resulting generative model becomes an optimal transport plan.
Although minimizing transportation costs has impacts beyond the context of generative modeling, we focus on accelerating the sampling of ODE-based generative models in this paper, so we design our method to achieve straight generative paths regardless of transportation costs.

\begin{figure}[h!]
    \centering
    \includegraphics[width=1\linewidth]{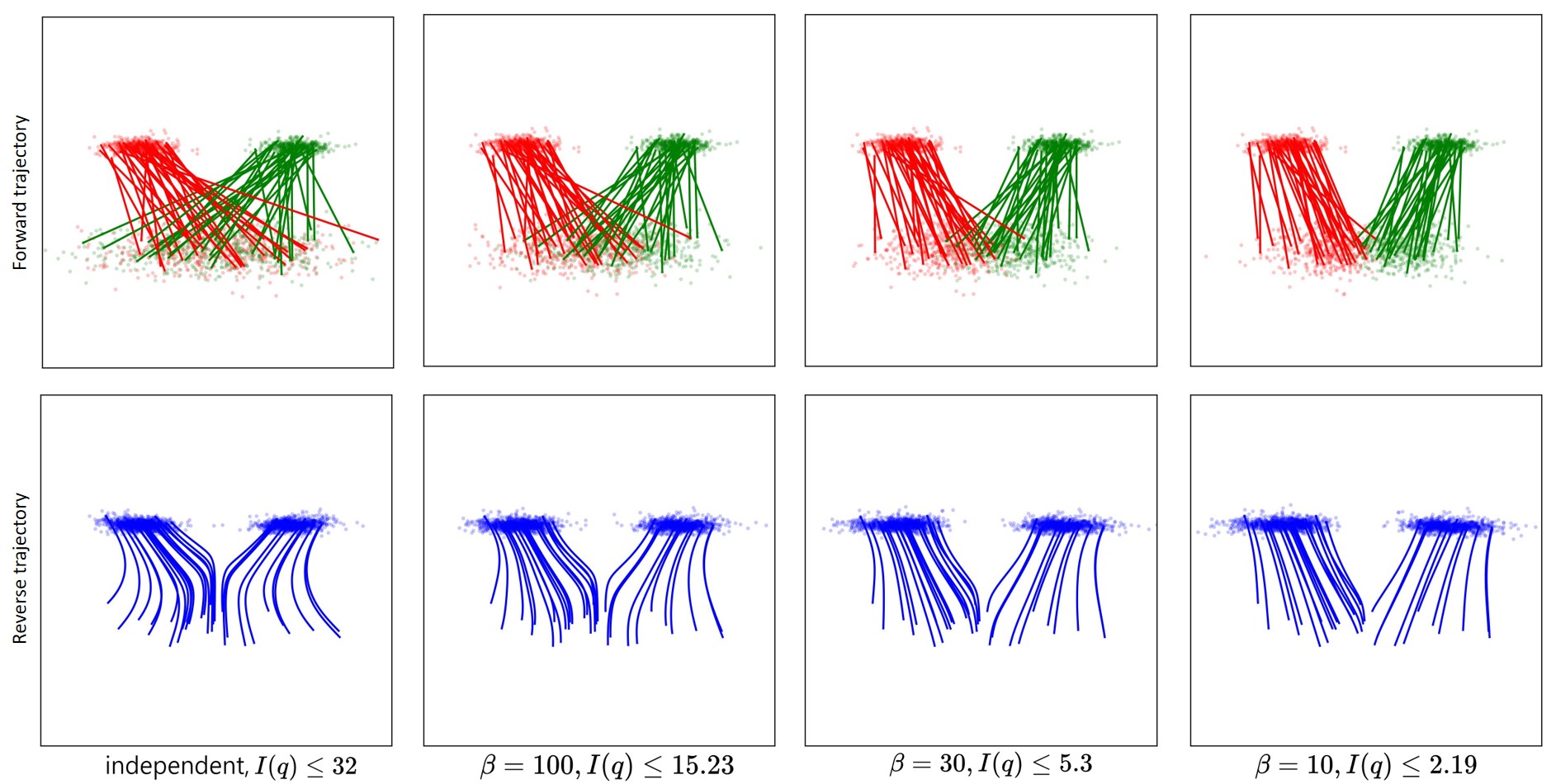} 
    \caption{The relationship between the degree of intersection between forward trajectories and curvature of reverse trajectories. The first column shows forward and reverse trajectories induced by the independent coupling $q(\x, \z) = p(\x)p(\z)$. As the degree of intersection between forward trajectories is decreased by lowering $\beta$, reverse paths are gradually straightened. }
    \label{fig:toy}
\end{figure}

\section{Experiment}
\subsection{2D dataset}
\label{subsection:image-generation}
Fig.~\ref{fig:toy} demonstrates the visual results and the estimated upper bounds of the degree of intersection on the 2D toy dataset. The leftmost column shows the forward trajectories induced by the independent coupling $q(\x, \z) = p(\x)p(\z)$ used in previous work. Since a data point can be mapped to any noise, the forward trajectories largely intersect with each other, and as a result, the reverse trajectories collapse toward the average direction where the actual density is low, and the curvature increases as the reverse trajectories need to bend toward the modes. As $\beta$ decreases, $q_\phi(\z | \x)$ tries to \textit{untie} the tangled trajectories, leading to low curvature.

\begin{figure}[h!]
    \centering
    \includegraphics[width=0.7\linewidth]{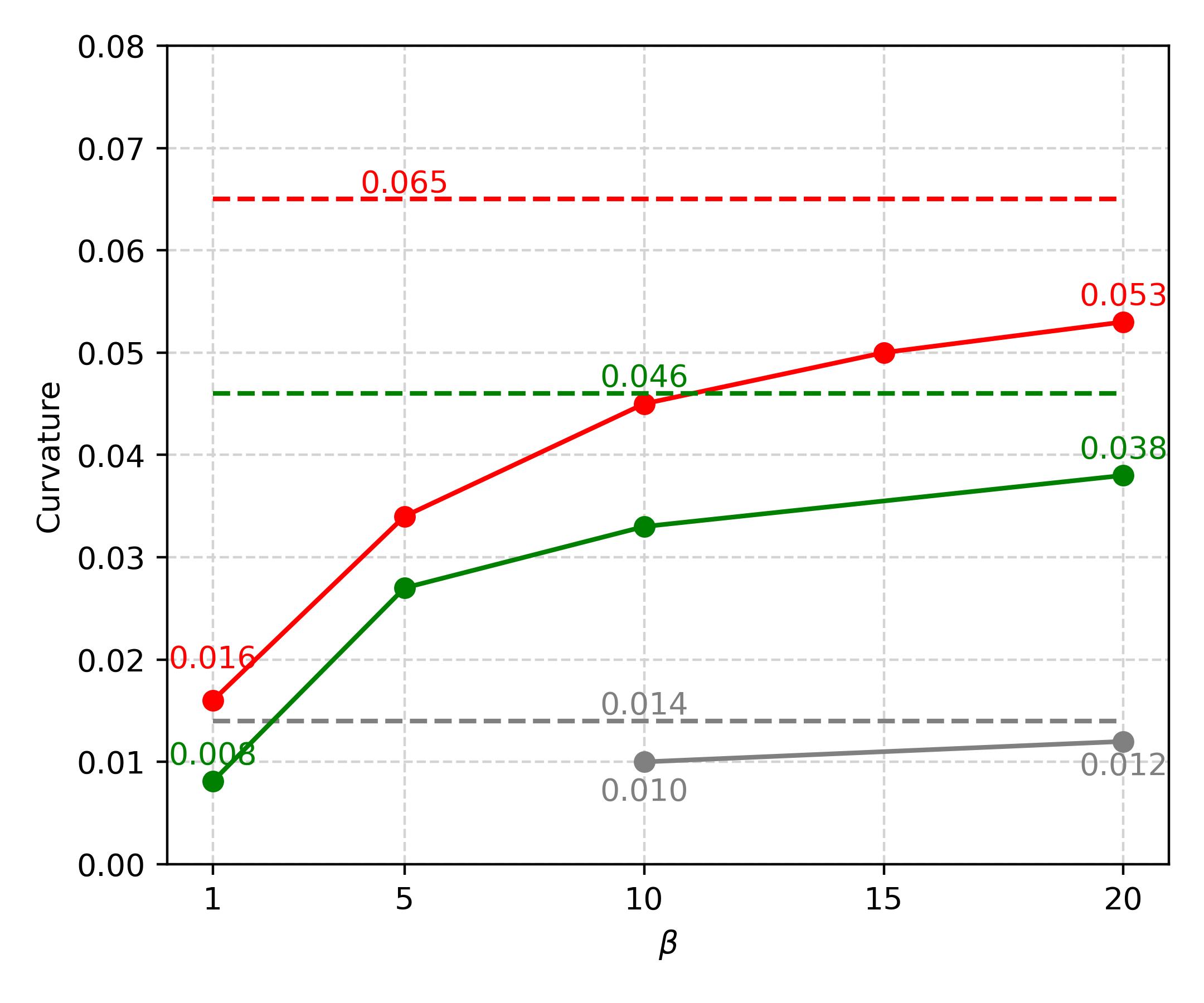}
    \vskip -0.15in
    \caption{Effects of $\beta$ on curvature. The results on MNIST, CIFAR-10, and CelebAHQ ($64 \times 64$) are indicated by \textcolor{red}{red}, \textcolor{teal}{green}, and \textcolor{gray}{gray} colors. Dashed lines indicate the curvatures of the independent coupling baselines.}
    \label{fig:curvature-graph}
\end{figure}

\begin{figure}[h!]
    \centering
    \includegraphics[width=0.8\linewidth]{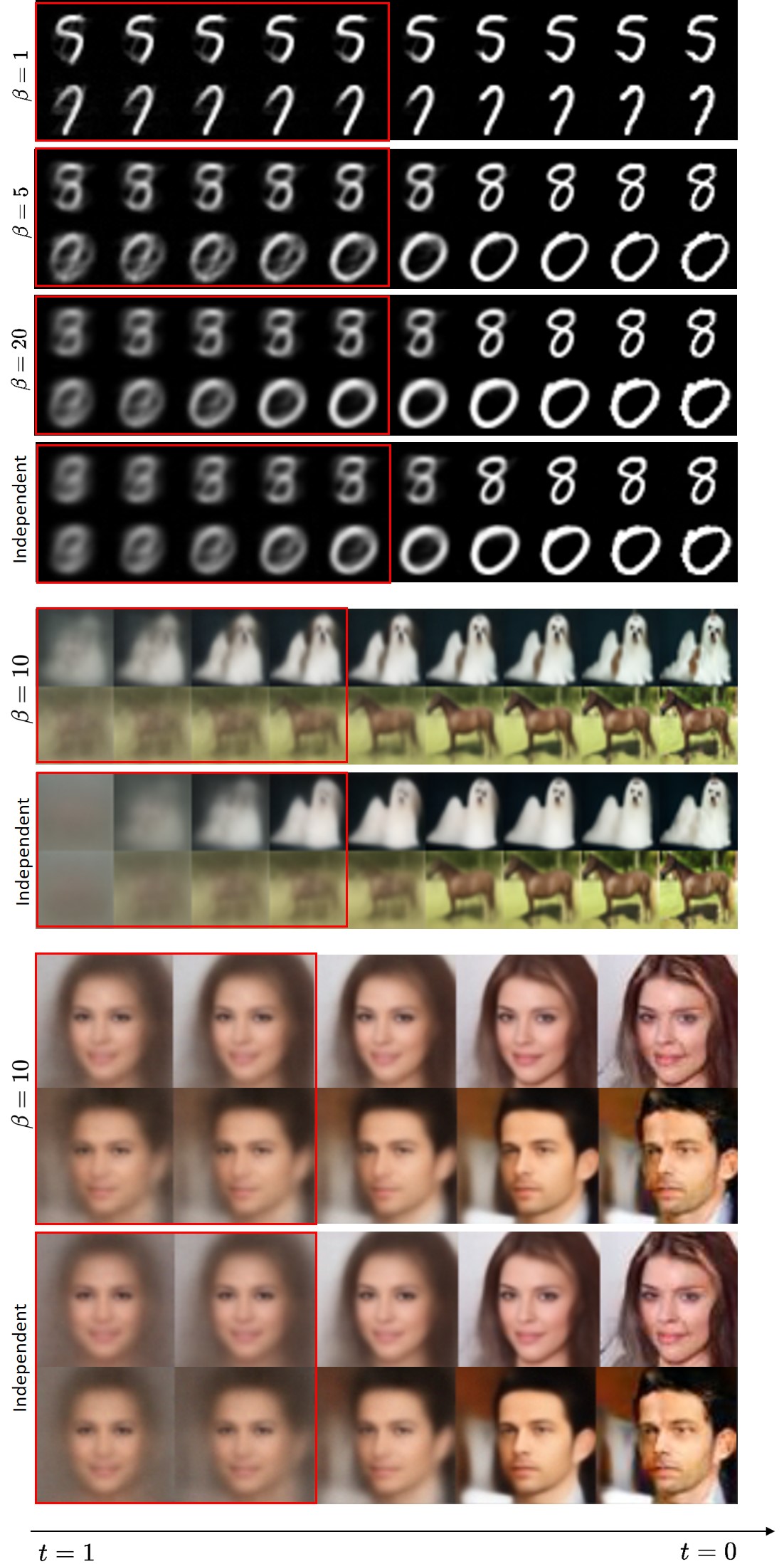}
    \vskip -0.15in
    \caption{Visualization of intermediate samples $\x_\thetab(\z_t, t)$ with varying $\beta$. Lower $\beta$ allows for sharper initial predictions, as indicated by red boxes.}
    \label{fig:curvature-vis}
    \vspace{-10pt}
\end{figure}

\subsection{Image generation}
\label{sec:image-generation}
We further conduct an experiment on the image dataset to investigate the relationship between $I(q)$ and $\mathbb E[C(\Z)]$ in the high-dimensional space. We estimate $\mathbb E[C(\Z)]$ by simulating $10,000$ generative trajectories using the Euler solver with 128 steps and then divide by the number of pixels.
As shown in Fig.~\ref{fig:curvature-graph}, the average curvature is the highest when using independent forward coupling $q(\x, \z) = p(\x)p(\z)$ and lowered as $\beta$ decreases. Fig.~\ref{fig:curvature-vis} shows that the generative vector field induced by independent coupling $q(\x, \z) = p(\x, \z)$ initially predicts the blurry images and then bends toward the mode, resulting in the high curvature. Lower $\beta$ yields more consistent results across the time steps.

\begin{figure}[h!]
\centering
\begin{subfigure}{0.54\linewidth}
\includegraphics[width=1\linewidth]{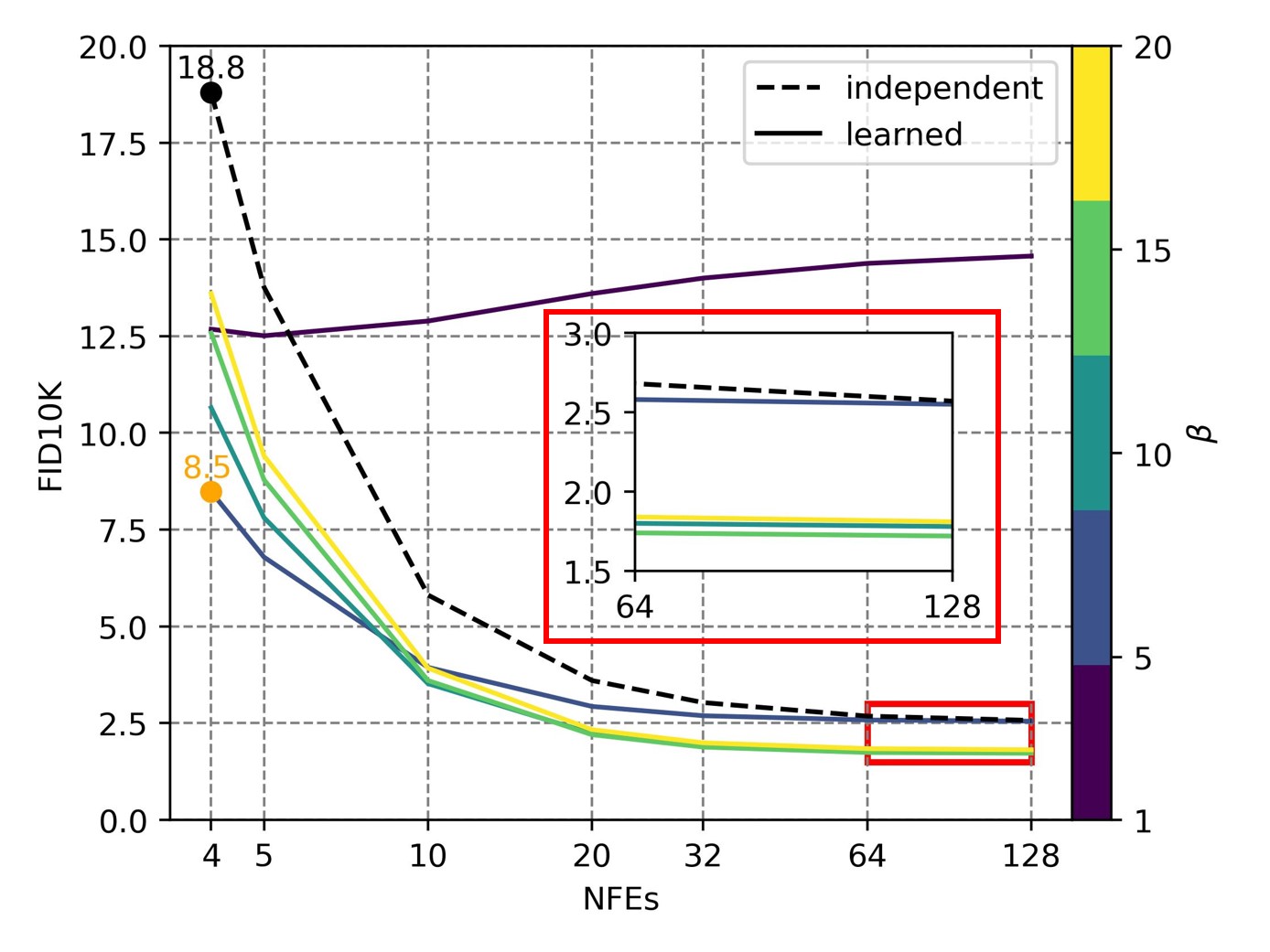}
\caption{Quantitative results}
\end{subfigure} 
\begin{subfigure}{0.45\linewidth}
\includegraphics[width=1\linewidth]{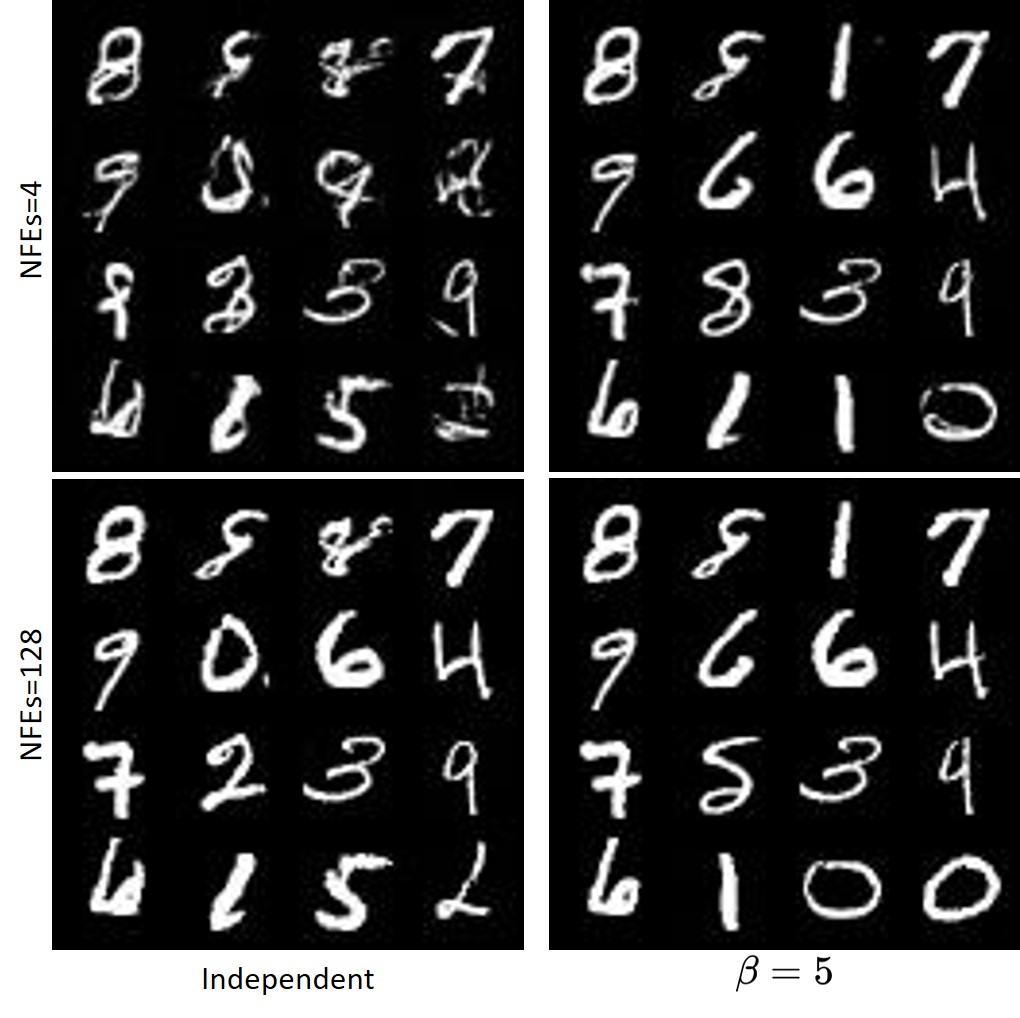}
\caption{Qualitative results}
\end{subfigure}
\caption{Trade-off between FID10K and the number of function evaluations (NFE) with varying $\beta$. The low curvature generative process produces more high-quality samples than the baseline with limited NFEs.}
\label{fig:graph}
\end{figure}

\begin{figure}[h!]
\centering
\begin{subfigure}{0.45\linewidth}
\includegraphics[width=1\linewidth]{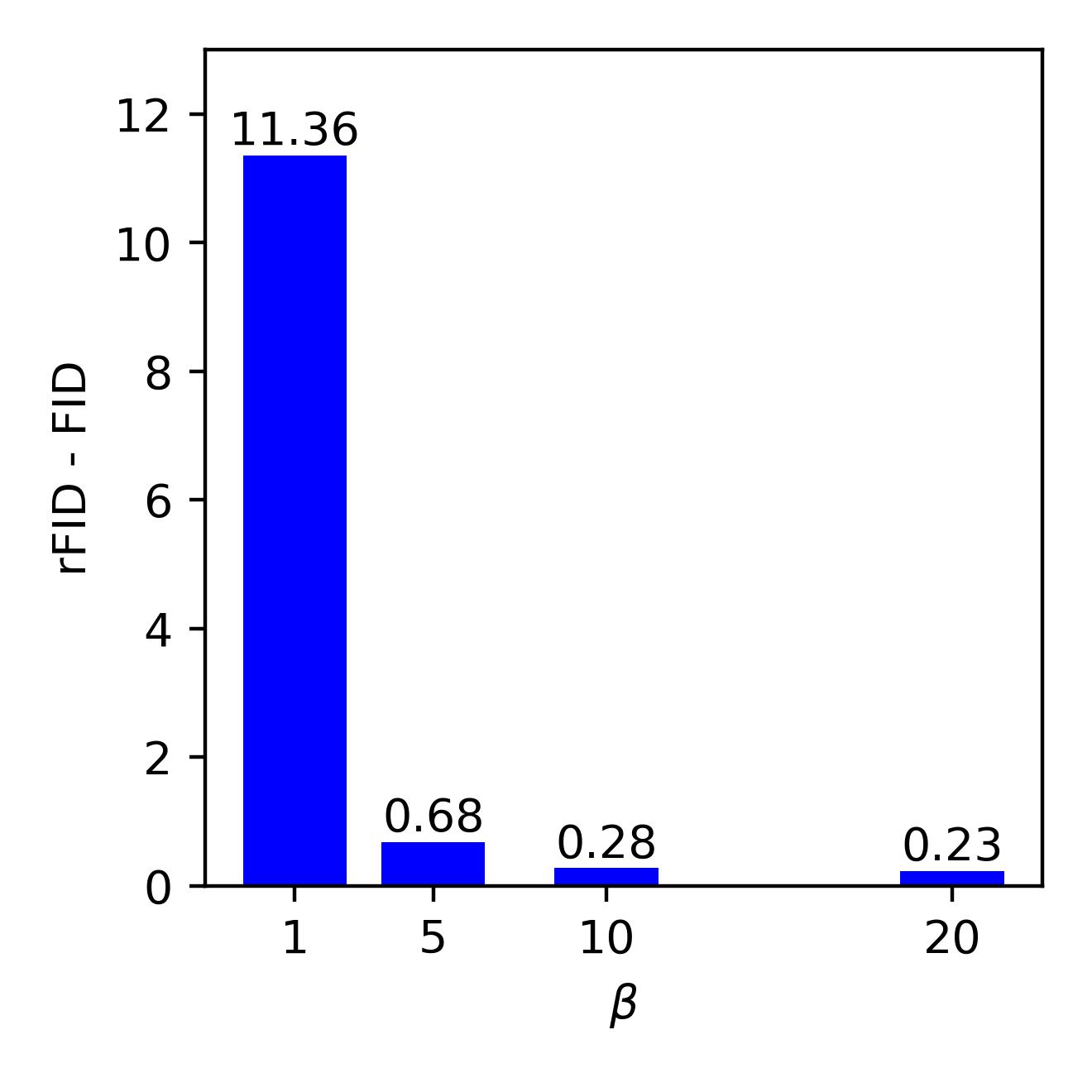}
\caption{FID gap with respect to $\beta$}
\end{subfigure} 
\begin{subfigure}{0.45\linewidth}
\includegraphics[width=1\linewidth]{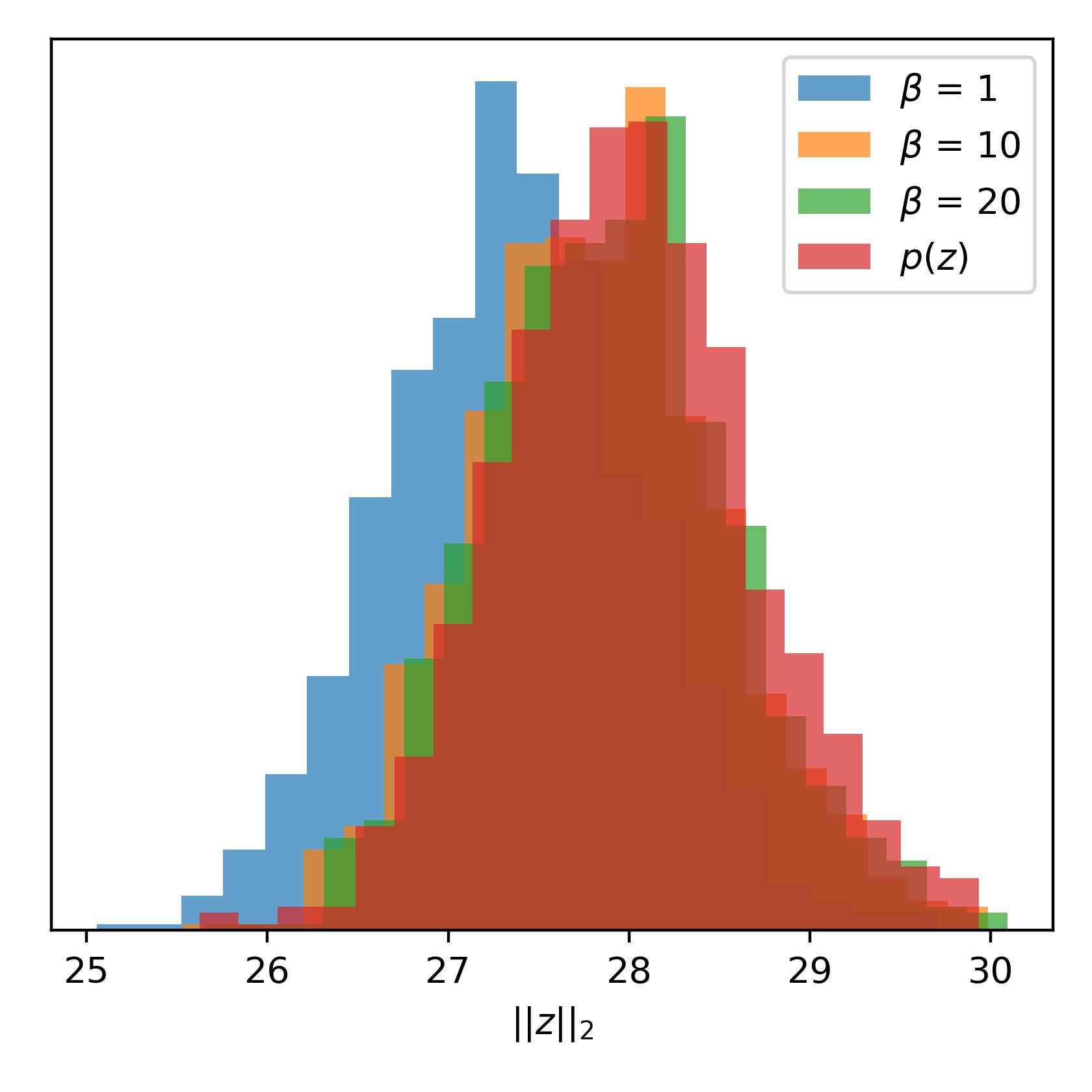}
\caption{Distribution of $||\z ||_2$}
\end{subfigure}
\caption{The gap between reconstruction FID (rFID) and FID values (a), and distribution of the norm of $\z \sim q_\phib(\z)$ (b). rFID is measured using samples reconstructed from $q_\phib(\z)$.}
\label{fig:prior-hole}
    \vspace*{-0.5cm}
\end{figure}

Fig.~\ref{fig:graph} shows that the model trained with the lower $\beta$ performs better with the limited NFEs and asymptotically approaches the performance of baseline indicated by a dashed line. When $\beta$ is as low as $1$, the generative process is almost straight, but the sample quality is degraded because of high $D_{KL}(q_\phi(\z) || p(\z))$ (i.e. the prior hole problem). As shown in Fig.~\ref{fig:prior-hole}, the gap between reconstruction FID (rFID) and FID is large when $\beta=1$ and gradually becomes smaller as $\beta$ increases. Moreover, the distribution of the norm of latent vectors gradually approaches $p(\z)$ as $\beta$ increases.
From this observation, we can see that $\beta$ is an important hyperparameter that determines the trade-off between sample quality and computational cost. We find that  there are little advantages of setting $\beta$ to $\infty$ as in previous work~\citep{liu2022flow,ho2020denoising}. It is an overkill for reducing the prior hole and leads to poor sampling efficiency.

\begin{figure*}[t!]
\centering
\includegraphics[width=1\linewidth]{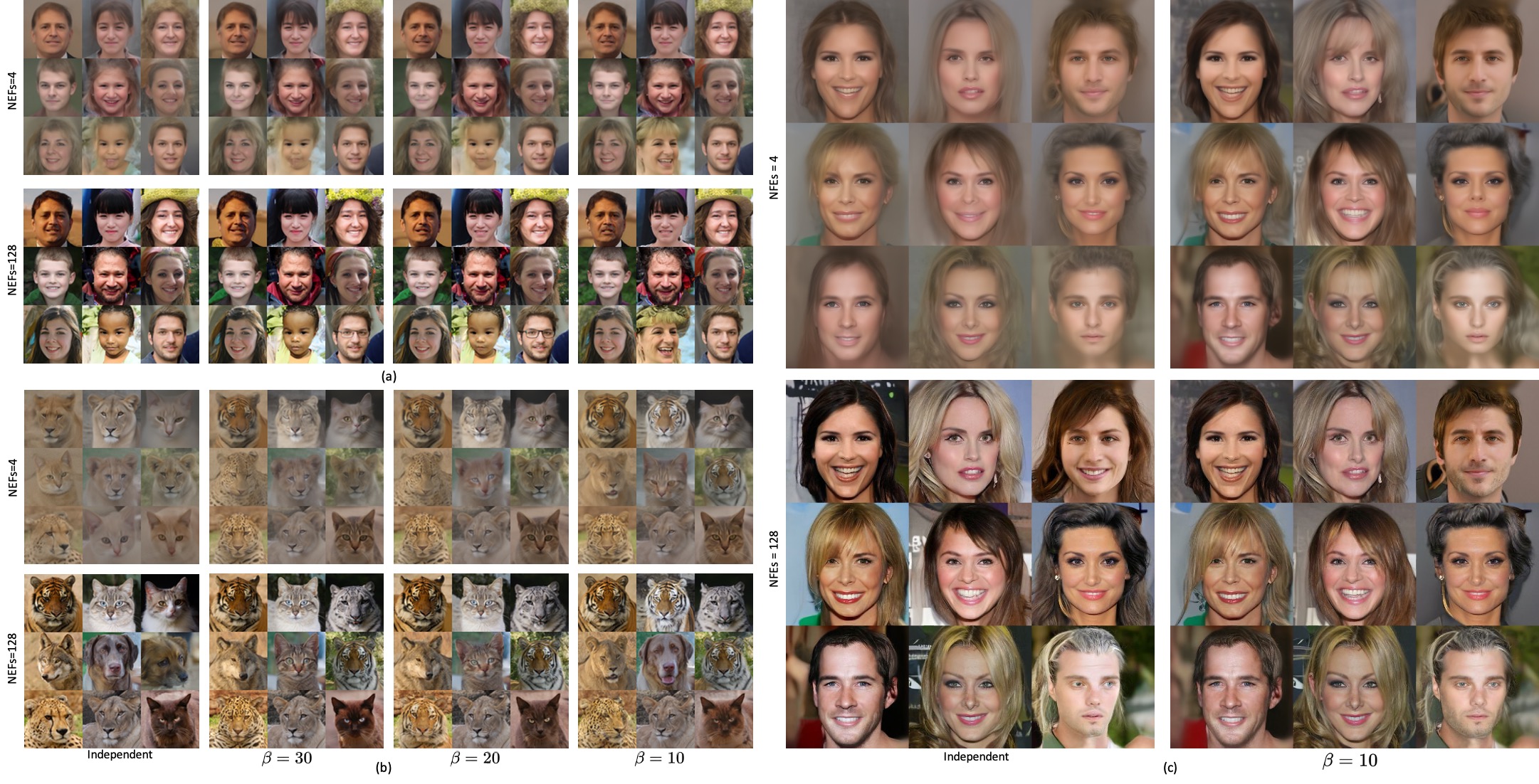}
    \vspace*{-0.7cm}
\caption{Qualitative comparison between our method and baseline on FFHQ 64 $\times$ 64 (a), AFHQ 64 $\times$ 64 (b), and CelebAHQ 256 $\times$ 256 (c) datasets.}
\label{fig:qualitative2}
\end{figure*}

In Fig.~\ref{fig:qualitative2} and Tab.~\ref{tab:quantitative}, we provide additional qualitative and quantitative comparisons between our method and rectified flow baseline on FFHQ 64 $\times$ 64, AFHQ 64 $\times$ 64, and CelebAHQ 256 $\times$ 256 datasets, which further confirm the validity of our method.

\begin{table*}
\begin{subtable}{1\textwidth}
\centering
\resizebox{0.5\textwidth}{!}{
\begin{tabular}{@{}llllllllll@{}}
\toprule
 & Setting \textbackslash \ NFEs & 4         & 5         & 10        & 20       & 32       & 64       & 128      &  \\ \midrule
 & $\beta = 10$                & \textbf{32.58} & \textbf{25.33} & \textbf{13.21} & 8.85     & 7.54     & 6.91     & 7.01     &  \\
 & $\beta = 20$                & 38.23     & 29.12     & 14.03     & 8.78     & 7.08     & 5.95     & 5.72     &  \\
 & $\beta = 30$                & 41.16     & 30.75     & 14.37     & \textbf{8.76} & \textbf{6.90} & \textbf{5.45} & \textbf{4.93} &  \\
 & Independent                 & 55.90     & 40.96     & 17.29     & 9.79     & 7.55     & 5.89     & 5.26     &  \\ \bottomrule
\end{tabular}
}
   \caption{FFHQ 64 $\times$ 64}\label{tab:sub_first}
\end{subtable}

\bigskip
\begin{subtable}{1\textwidth}
\centering

\resizebox{0.5\textwidth}{!}{
\begin{tabular}{@{}llllllllll@{}}
\toprule
 & Setting \textbackslash \ NFEs & 4         & 5         & 10        & 20       & 32       & 64       & 128      &  \\ \midrule
 & $\beta = 10$                & \textbf{21.80} & \textbf{18.04} & 11.80     & 9.05     & 8.22     & 7.47     & 7.21     &  \\
 & $\beta = 20$                & 25.73     & 20.11     & \textbf{10.56} & 6.89     & 5.74     & 4.92     & 4.55     &  \\
 & $\beta = 30$                & 30.84     & 23.08     & 11.17     & \textbf{6.66} & \textbf{5.37} & \textbf{4.40} & \textbf{3.96} &  \\
 & Independent                 & 54.10     & 42.64     & 18.53     & 8.60     & 6.19     & 4.85     & 4.36     &  \\ \bottomrule
\end{tabular}
}
   \caption{AFHQ 64 $\times$ 64}\label{tab:sub_second}
\end{subtable}

\bigskip
\begin{subtable}{1\textwidth}
\centering
\resizebox{0.5\textwidth}{!}{
\begin{tabular}{@{}llllllllll@{}}
\toprule
 & Setting \textbackslash \ NFEs & 4         & 5         & 10        & 20        & 32        & 64        & 128       &  \\ \midrule
 & $\beta = 10$                & \textbf{58.30} & \textbf{51.02} & 33.53     & 22.91 & 19.49     & 17.57     & 16.94     &  \\
 & $\beta = 40$                & 62.21     & 52.92     & \textbf{31.70} & \textbf{18.70} & \textbf{14.03} & \textbf{11.39} & \textbf{10.37} &  \\
 & Independent                 & 100.39    & 84.50     & 48.95     & 26.42     & 18.45 & 12.78 & 10.38     &  \\ \bottomrule
\end{tabular}
}
   \caption{CelebAHQ 256 $\times$ 256}
   \label{tab:sub_third}
\end{subtable}

\caption{FID10K comparison on three image datasets.}
\label{tab:quantitative}
\end{table*}

\subsection{Distillation}
\begin{figure}[h!]
    \centering
    \begin{subfigure}{0.4\linewidth}
    \includegraphics[width=1\linewidth]{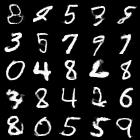}
    \caption{$\beta=10$}
    \end{subfigure} 
    \begin{subfigure}{0.4\linewidth}
    \includegraphics[width=1\linewidth]{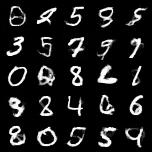}
    \caption{Independent}
    \end{subfigure}
    \vskip -0.1in
    \caption{Synthesis results of one-step models.}
    \label{fig:distillation}
        \vspace*{-0.5cm}
\end{figure}

Even though distillation is an effective way to train the one-step student models from the teacher diffusion models, the performance of the student model is suboptimal due to the distillation error~\citep{liu2022flow,luhman2021knowledge,salimans2022progressive}. Given that the teacher trajectories with higher NFEs are more difficult to distill, our low-curvature generative ODEs would make less distillation error since they achieve the same level of sample quality using relatively lower NFEs. Based on this intuition, we investigate the effect of our method on reducing the distillation error. As shown in Table~\ref{tab:distillation}, the teacher ODE with $\beta=10$ 
achieves a similar FID score using half as many NFEs compared to the baseline model. This resulted in a smaller distillation error and an improved FID score of the one-step model while reducing the cost of generating paired data by half. 
Fig.~\ref{fig:distillation} demonstrates that our method with $\beta=10$ obtains a superior one-step model than baseline with independent coupling in terms of sample fidelity.

\begin{table}[h!]
\caption{Effects of curvature on distillation performance. The results of independent coupling and our method with $\beta=10$ are reported. Distillation error is measured as a mean-squared error on the test set.}
\vskip 0.15in
\small

\centering
\resizebox{0.4\textwidth}{!}{
\begin{tabular}{@{}ccccc@{}}
\toprule
                            & \multicolumn{2}{c}{Independent} & \multicolumn{2}{c}{$\beta = 10$} \\ \midrule
                            & FID / Error         & NFEs      & FID / Error              & NFEs  \\ \midrule
\multicolumn{1}{l}{Teacher} & 3.60 / -            & 20        & 3.52 / -                 & 10    \\
Distilled                   & 6.25 / 0.0208       & 1         & \textbf{4.41 / 0.0157}   & 1     \\ \bottomrule
\end{tabular}}
\label{tab:distillation}
\end{table}

\subsection{Size of encoder}
Since we train $q_\phib(\z | \x)$, a natural question is how much additional computational cost is needed for training our model. We experiment with two settings of the encoder, \texttt{same} and \texttt{small}. In \texttt {same} setting, we use the identical architecture with a generative component except that the number of output channels is twice for predicting a diagonal covariance. In \texttt{small} setting, we use roughly 20 times smaller architecture for the encoder model. See Appendix~\ref{appendix:implementation} for a detailed configuration. As shown in Table~\ref{tab:encoder}, a small encoder performs just as well or even better than a larger encoder, and part of the reason is that we can use a larger batch size in the \texttt{small} setting. The additional cost of our method is negligible with the use of a lightweight architecture for $q_\phib(\z | \x)$, so we stick to \texttt{small} setting throughout our experiments.

\begin{table}[h]
\caption{Performance comparison of \texttt{same} and \texttt{small} encoder settings on CIFAR-10 dataset, measured by FID10K.}
\vskip 0.15in
\small
\centering
\resizebox{0.46\textwidth}{!}{
\begin{tabular}{lclllll}
\toprule
Setting \textbackslash \ NFEs
& 128                            & \multicolumn{1}{c}{40}         & \multicolumn{1}{c}{20}         & \multicolumn{1}{c}{10}          & \multicolumn{1}{c}{5}           & \multicolumn{1}{c}{4}           \\ \cmidrule{1-7}
Same Encoder   & 5.52                           & 6.23                           & 7.74                           & 11.49                           & 22.90                           & 30.97                           \\
Small Encoder  & \textbf{5.39} & \textbf{6.07} & \textbf{7.51} & \textbf{11.19} & \textbf{22.33} & \textbf{30.16} \\ \bottomrule
\end{tabular}}
\label{tab:encoder}
\end{table}

\begin{table*}[t!]
\centering
\caption{Comparison with state-of-the-arts on CIFAR-10 dataset. * Our reimplementation.}
\vskip 0.15in
\small
\resizebox{0.7\textwidth}{!}{
\begin{tabular}{lcccc}
        \toprule
        Method & NFEs($\downarrow$) & IS ($\uparrow$) & FID ($\downarrow$) & Recall ($\uparrow$) \\ \cmidrule{1-5}
        \emph{GANs} & 
        \multicolumn{4}{l}{} \\ 
        \cmidrule{1-5}           
        StyleGAN2~\citep{karras2020training} & 1 & 9.18 & 8.32 & 0.41 \\
         StyleGAN2 + ADA~\citep{karras2020training} & 1 & 9.40 & 2.92 & 0.49 \\
         StyleGAN2 + DiffAug~\citep{zhao2020differentiable} & 1 & 9.40 & 5.79 & 0.42 \\
         \cmidrule{1-5}   
         \emph{ODE/SDE-based models} & \multicolumn{4}{l}{} \\  
         \cmidrule{1-5}   
         Denoising Diffusion GAN (T=1)~\citep{xiao2021tackling} & 1 & 8.93 & 14.6 & 0.19 \\ 
        DDPM~\citep{ho2020denoising} & 1000 & 9.46 & 3.21 & 0.57 \\
        NCSN++ (VE SDE)~\citep{song2020score} & 2000 & 9.83 & 2.38 & 0.59  \\
        LSGM~\citep{vahdat2021score} & 138 & - & 2.10 & - \\
        DFNO~\citep{zheng2022fast} & 1 & - & 5.92 & -\\
        Knowledge distillation~\citep{luhman2021knowledge} & 1 & 8.36 & 9.36 & 0.51 \\
        Progressive distillation~\citep{salimans2022progressive} & 1 & - & 9.12 & - \\
         Rectified Flow (RK45)~\citep{liu2022flow} & 127 & {9.60} & {2.58} & {0.57} \\
         {2-Rectified Flow (RK45)} & 110 & 9.24 & 3.36 & 0.54 \\
         {3-Rectified Flow (RK45)} & 104 & 9.01 & 3.96 & 0.53 \\
         2-Rectified Flow Distillation & 1 & 9.01 & 4.85 & 0.50 \\
         \cmidrule{1-5}   
         \emph{Our results} & \multicolumn{4}{l}{} \\
         \cmidrule{1-5}   
         Rectified Flow* (config A, RK45) & 134 & 9.18 & 2.87 & - \\
         Rectified Flow* (config B, RK45) & 132 & 9.48  & 2.66 & 0.62 \\
         Rectified Flow* (config B, \heun)  & 9 & 8.48 & 12.92  & - \\
         Rectified Flow* (config B, \heun) & 5 & 7.04 & 37.19 & - \\
         Ours ($\beta=20$, config B, RK45) & 118 & 9.55 & 2.45 & 0.64 \\
         Ours ($\beta=20$, config B, \heun) & 9 & 8.75 & 9.96 & - \\
         Ours ($\beta=20$, config B, \heun) & 5 & 7.83 & 24.40 & - \\
         Ours ($\beta=10$, config A, RK45) & 110 & 9.32 & 3.37& 0.61 \\
         Ours ($\beta=10$, config A, \heun) & 9 & 8.67 & 8.66 & - \\
         Ours ($\beta=10$, config A, \heun) & 5 & 8.09 & 18.74& -\\

\bottomrule 
         
    \end{tabular}
    }
\label{tab:sota}
\end{table*}

\subsection{Comparison with state-of-the-arts}

Table~\ref{tab:sota} shows the unconditional synthesis results of our approach on the CIFAR-10 dataset. Results of recent methods are also provided as a reference. We experiment with two configurations, config A and config B, which we detail in Appendix~\ref{appendix:implementation}. We try three solvers, Euler solver, \heun, and the black-box RK45 method from Scipy~\citep{virtanen2020scipy}, and find that RK45 works well when we are able to fully simulate ODEs while \heun \ performs better than other solvers with small NFEs.
As shown in the table, we can see that the performance gap between our method and the baseline is huge when the sampling budget is limited. For instance, our method with $\beta=10$ achieved an FID score of 18.74, which is significantly better than the baseline's score of 37.19 when NFEs is 5. Surprisingly, our method with $\beta=20$ exhibits superior sample qualities across all NFEs, even in the case of full sampling using the RK45 solver. See Fig.~\ref{fig:qualitative} for visual comparison. Additional qualitative results are provided in Appendix~\ref{appendix:additional}.

\begin{figure}[h!]
\centering
\includegraphics[width=0.95\linewidth]{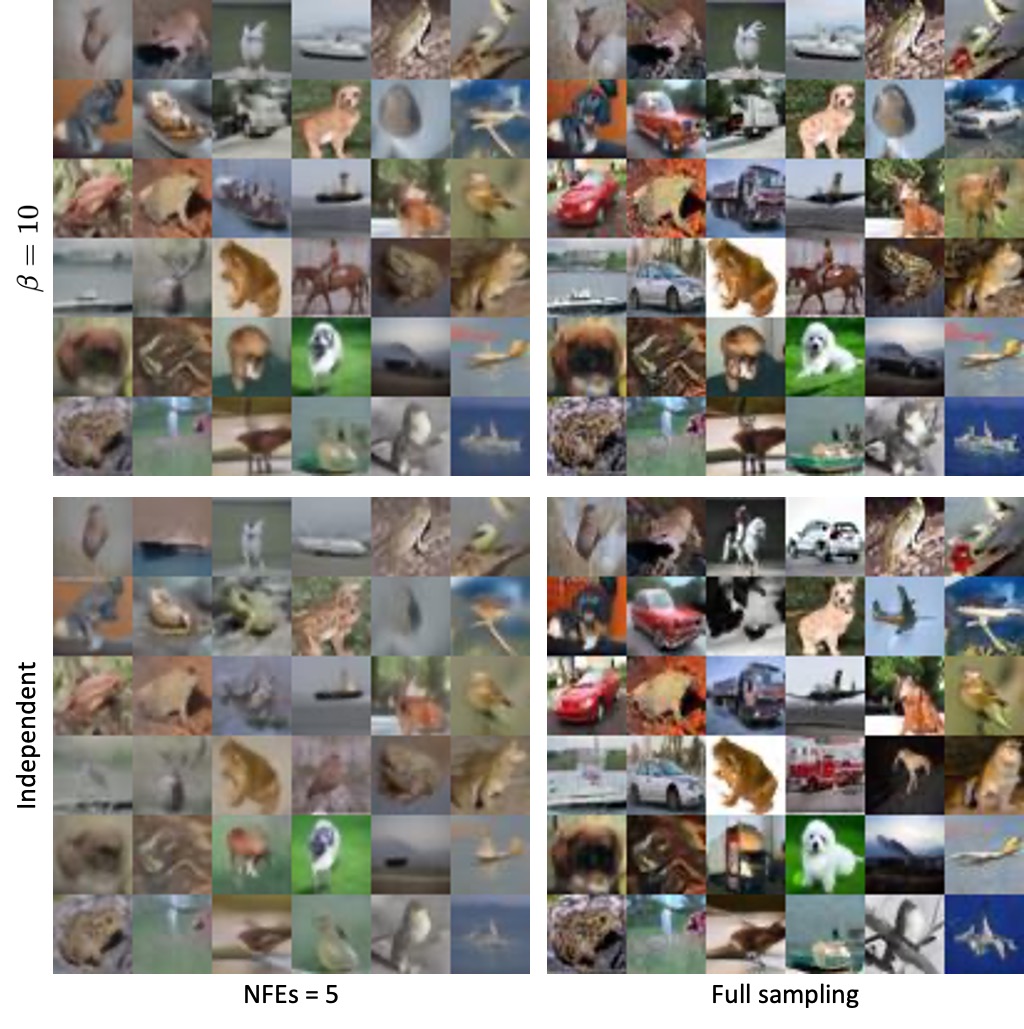}
    \vspace*{-0.5cm}
\caption{Qualitative comparison between our method ($\beta=10$) and baseline on CIFAR-10.}
\label{fig:qualitative}
\end{figure}

\section{Discussion and Limitations}
One limitation of our method is that for our encoding distribution $q_\phib(\z | \x)$, we use a Gaussian distribution for theoretical and practical conveniences: sampling is easily implemented in a differentiable manner, and KL divergence is tractable. However, our simple Gaussian encoder cannot eliminate the intersection completely. We believe that it would be beneficial to use a more flexible encoding distribution, for instance, using the hierarchical latent variable as in \citet{child2020very,vahdat2020nvae}.

Additionally, the trade-off between sample quality and computational cost is determined by the value of $\beta$, which must be manually selected by a practitioner. In Sec.~\ref{sec:image-generation}, we observe that too small $\beta$ value causes the prior hole problem.
This is problematic as one has to train a model from scratch for each value of $\beta$, which would potentially lead to excessive energy consumption. However, using a reasonably high value of $\beta$ consistently outperforms the baseline regardless of the sampling budget, as shown in our experiments. Therefore, one could reduce the sampling cost without compromising performance by conservatively setting $\beta$ to a high value in most cases.

\section{Conclusion}
In this paper, we mainly discussed the curvature of the ODE-based generative models, which is crucial for sampling efficiency. We revealed the relationship between the degree of intersection between forward trajectories and the curvature and presented an efficient algorithm to reduce the intersection by training a forward coupling. We demonstrated that our method successfully reduces the trajectory curvature, thereby enabling accurate ODE simulation with significantly less sampling budget. Furthermore, we showed our method effectively decreases the distillation error, improving the performance of one-step student models. Our approach is unique and complementary to other acceleration methods, and we believe it can be used in conjunction with other techniques to further decrease the sampling cost of ODE-based generative models.

\section{Societal Impacts.}
We anticipate this work will have positive effects, as our method reduces the computational costs required during the sampling of ODE-based generative models. However, the same technology can also be used to create malicious content, and thus, proper regulations need to be put in place to ensure that this technology is used responsibly and ethically.

\section*{Acknowledgements}
This work was supported by the National Research Foundation of Korea under Grant NRF-2020R1A2B5B03001980, by the Korea Medical Device Development Fund grant funded by the Korea government (the Ministry of Science and ICT, the Ministry of Trade, Industry and Energy, the Ministry of Health \& Welfare, the Ministry of Food and Drug Safety) (Project Number: 1711137899, KMDF\_PR\_20200901\_0015), and Field-oriented Technology Development Project for Customs Administration through National Research Foundation of Korea funded by the Ministry of Science \& ICT and Korea Customs Service (NRF-2021M3I1A1097938), by Institute of Information \& communications Technology Planning \& Evaluation (IITP) grant funded by the Korea government (MSIT, Ministry of Science and ICT) (No. 2022-0-00984, Development of Artificial Intelligence Technology for Personalized Plug-and-Play Explanation and Verification of Explanation), (No.2019-0-00075, Artificial Intelligence Graduate School Program), and ITRC (Information Technology Research Center) support program (IITP-2021-2020-0-01461). This work was also 
supported by  the KAIST Key Research Institute (Interdisciplinary Research Group) Project.

\bibliography{reference}
\bibliographystyle{icml2023}

\newpage
\appendix
\onecolumn
\section{Preliminaries}
\label{appendix:preliminaries}
\subsection{Rectified flows}
Diffusion models have been interpreted as the variational approaches~\citep{sohl2015deep,ho2020denoising} or score-based models~\citep{song2019generative,song2020score}, and their deterministic samplers are derived post hoc. However, stochasticity is not a key factor in the success of these models. The state-of-the-art performance can be achieved without stochasticity~\citep{karras2022elucidating}, and the incorporation of stochasticity makes sampling slow and complicates the theoretical understanding.
Rectified flow~\citep{liu2022flow} provides a useful viewpoint for explaining the recent iterative methods~\citep{ho2020denoising,song2019generative} from a pure ODE perspective. For the purpose of brevity, we only consider the variance-preserving diffusion models~\citep{song2020score} and 1-Rectified Flow here. Readers are encouraged to refer to \citet{liu2022flow,liu2022rectified} for a more comprehensive explanation. 

 The variance-preserving diffusion models define the following noise distribution
\begin{align}
    q(\x_t | \x) = \mathcal N(\alpha(t)\x, (1- \alpha(t)^2) \I),
\end{align}
where $\alpha(t)$ is set to $\exp(-\frac{1}{2}\int_0^t (as + b) \ ds)$ with $a = 19.9$ and $b = 0.1$. From a rectified flow (or similarly, stochastic interpolant~\citep{albergo2022building,albergo2023stochastic}) perspective, another way to see this is to consider the nonlinear interpolation between $\x$ and $\z$ sampled independently from $q(\x, \z) = p(\x)p(\z)$:
\begin{align}
    \x_t(\x, \z) = \alpha(t)\x + \sqrt{1- \alpha(t)^2}\z
\end{align}
This \textit{forward flow} represents the dynamics of particles that move from $p(\x)$ to $p(\z)$. Note that this cannot be used for generative modeling as it requires $\x$ to compute the velocity. To estimate the velocity without having $\x$, a neural network $\x_\theta(\x_t, t)$ is trained by optimizing
\begin{align}
    \min_\thetab \mathbb E_{\x, \z, t}[\lambda(t)||\x - \x_\theta(\x_t(\x, \z),t)||_2^2].
\end{align}

From this viewpoint, the choice of nonlinear interpolation is unnatural since it unnecessarily increases the curvature of both forward and reverse (generative) trajectories. For this reason, \citet{liu2022flow} defines the following constant-velocity flow with an initial value $\x$ and endpoint $\z$:
\begin{align}
    d\x_t (\x, \z) &= (\z - \x)dt \\
    \x_0(\x, \z) &= \x
\end{align}
Instead of predicting $\x$, they directly train a vector field $\mathbf v_\theta(\x_t, t)$ to match the velocity of forward flow by minimizing the following loss
\begin{equation}
    L_{FM} = \int_0^1 \mathbb E[|| (\z - \x) - \vb_\theta(\x_t, t) ||_2^2]\ dt,
    \label{eq:appendix:flow-matching}
\end{equation}
where $\vb_\theta(\x_t,t) = \mathbb E[\z - \x | \x_t]$ in the optima. Samples are drawn by solving the following ODE backward:
\begin{equation}
    d\z_t = \vb_\theta(\z_t,t)dt
    \label{eq:appendix:flow-sampling}
\end{equation}
It is shown that Eq.~\eqref{eq:appendix:flow-sampling} yields the same marginal distribution as the forward flow at every $t$ (see Theorem 3.3 in \citet{liu2022flow}).

Given $\x_t = (1-t)\x + t\z$ and $\z - \x = (\x_t - \x)/t$, we can further find the connection with diffusion models by reparameterizing $\vb_\theta(\x_t,t) = (\x_t - \x_\theta(\x_t, t)) / t$ and writing Eq.~\eqref{eq:appendix:flow-matching} as
\begin{align}
    \int_0^1 \mathbb E[|| (\z - \x) - \vb_\theta(\x_t, t) ||_2^2]\ dt
    &= \int_0^1 \mathbb E[|| (\x_t - \x)/t - \vb_\theta(\x_t, t) ||_2^2]\ dt
    \\
    &= \int_0^1 \mathbb E[|| (\x_t - \x)/t - (\x_t - \x_\theta(\x_t, t))/t ||_2^2]\ dt
    \\
    &= \int_0^1 \mathbb E[\frac{1}{t^2}|| \x - \x_\theta(\x_t, t)) ||_2^2]\ dt.
    \label{appendix:eq:fm-loss}
\end{align}
This is equivalent to Eq.~\eqref{eq:dsm} with $\lambda(t) = 1/t^2$, and Eq.~\eqref{eq:appendix:flow-sampling} is equal to Eq.~\eqref{eq:flow-sampling}. To our knowledge, the effectiveness of Eq.~\eqref{eq:appendix:flow-sampling} in reducing the truncation error is first examined in \citet{karras2022elucidating} under the variance-exploding scheme. 
\subsection{Flow matching}
\label{appendix:flow-matching}
Recently, \citet{lipman2022flow} proposed the flow matching method to learn CNFs in a simulation-free manner. In this section, we show the similarity between flow matching and rectified flows. We borrow notations from \citet{lipman2022flow} here. \citet{lipman2022flow} define a time-conditional probability distribution $p_t(x)$ where $p_0(x)$ is a Gaussian distribution and $p_1(x)$ is a data distribution. They further define a conditional distribution
\begin{align}
    p_t(x|x_1) &= \mathcal N(x | \mu_t(x_1), \sigma_t(x_1)^2I) \\
    \mu_t(x) &= tx_1 \text{ and}\  \sigma_t(x) = 1 - t \\
    u_t(x|x_1) &= \frac{x_1 - x}{1 - t}
\end{align}
for their OT-VFs formulation. Note that we set $\sigma_{min}$ to $0$ here. Then, the training loss is 
\begin{align}
    \min_{\theta} \mathbb E_{t, p_1(x_1), p_t(x|x_1) }||v_t(x;\theta) - u_t(x|x_1)||^2 \\
    = \mathbb E \left\|v_t(x;\theta) - \frac{x_1 - x}{1-t} \right\|^2.
\end{align}
To generate samples, they solve the following ODE:
\begin{align}
    d\phi_t(x) &= v_t(\phi_t(x); \theta)dt \\
    \phi_0(x)  &= x, x \sim p_0
\end{align}
Since \citet{lipman2022flow} define $p_1(x)$ as a data distribution and $p_0(x)$ as a standard normal distribution in contrast to our work, we do the following substitutions for comparison:

\begin{align}
    t &\rightarrow 1-s \\
    dt &\rightarrow -ds \\
    \phi_t(x) &\rightarrow \z_s(x) \\
    v_{t}(x; \theta) &\rightarrow -\boldsymbol {v}_\thetab (x, s) \\
    x_1 &\rightarrow \x \\
\end{align}
As a result, we have
\begin{align}
    u_t(x|x_1) &= \frac{x_1 - x}{1 - t} = \frac{\x - x}{s}.
\end{align}

and the following training loss:
\begin{align}
    \min_\theta \mathbb E \left\|\boldsymbol {v}_\thetab (x, s) - \frac{x - \x}{s}\right\|^2
\end{align}

Replacing $x$ with $ \x_s = (1-s)\x + s\z$ for $\z \sim \mathcal N(0, I)$, the training loss becomes
\begin{align}
    \min_\theta \mathbb E ||\boldsymbol {v}_\thetab (\x_s, s) -  (\z - \x)||^2,
\end{align}
and the generative ODE becomes
\begin{align}
    d\z_s (\z)= \boldsymbol v_\thetab(\z_s(\z),s)ds
\end{align}
which are equivalent to Eqs.~\eqref{eq:appendix:flow-matching} and \eqref{eq:appendix:flow-sampling}.

\section{Derivation of Loss Function}
\label{appendix:loss}
\subsection{Estimating $D_{KL}(q_\phi(\z) || p(\z))$}
\label{appendix:kl}

We factorize $D_{KL}(q_\phib(\z) || p(\z))$ via following algebraic manipulation:
\begin{align}
    D_{KL}(q_\phib(\z) || p(\z)) &= 
    \mathbb E_{p(\x)}\mathbb E_{q_\phib(\z | \x)}\left[\log \frac{q_\phib(\z)}{p(\z)}\right]
    \\
    &= \mathbb E_{p(\x)}\mathbb E_{q_\phib(\z | \x)} \left[\log \frac{q_\phib(\z)}{q_\phib(\z|\x)} + \log \frac{q_\phib(\z|\x)}{p(\z)}\right] \\
    &= \mathbb E_{p(\x)}\mathbb E_{q_\phib(\z | \x)} \left[\log \frac{q_\phib(\z)p(\x)}{q_\phib(\z|\x)p(\x)}\right] + \mathbb E_{p(\x)}[D_{KL}(q_\phib(\z | \x) || p(\z))] \\
    &= - \mathbb I_{q_\phib(\x, \z)}(\x, \z) + \mathbb E_{p(\x)}[D_{KL}(q_\phib(\z | \x) || p(\z))]
\end{align}

We can derive the variational lower bound of the mutual information as
\begin{align}
    \mathbb I_{q_\phib(\x, \z)}(\x, \z) &= H(\x) - H(\x | \z) \\
    &= H(\x) + \mathbb E_{q_\phib(\x, \z)} [\log q_\phib(\x | \z)] \\
    &= H(\x) + \mathbb E_{q_\phib(\x, \z)} \left[\log p_\psib(\x | \z) + \log \frac{q_\phib(\x | \z)}{p_\psib(\x | \z)}\right] \\
    &= H(\x) + \mathbb E_{q_\phib(\x, \z)} [\log p_\psib(\x | \z) ] + \mathbb E_{q_\phib(\z)}[D_{KL}(q_\phib(\x | \z) || p_\psib(\x | \z))] \\
    &\geq H(\x) + \mathbb E_{q_\phib(\x, \z)} [\log p_\psib(\x | \z) ],
\end{align}
where the bound is tight when the variational distribution $p_\psi (\x | \z)$ is equal to $q_\phib(\x | \z)$. For that, we need to optimize $\min_\psib \mathbb E_{q_\phib(\z)} [-\log p_\psib(\x | \z)]$, which becomes the reconstruction loss $\mathbb E_{p(\x)} \mathbb E_{q_\phib(\z | \x)}\left[\frac{||x_\psib(\z) - \x||_2^2}{2\sigma^2}\right]$ if we set $p_\psib(\x | \z) = \mathcal N(\x; \x_\psib(\z), \sigma^2 \mathbf I).$ Consequently, we arrive at
\begin{align}
    D_{KL}(q_\phib(\z) || p(\z)) \leq \inf_\psib \ \mathbb E_{p(\x)} \mathbb E_{q_\phib(\z | \x)}\left[\frac{||x_\psib(\z) - \x||_2^2}{2\sigma^2}\right] + \mathbb E_{p(\x)}[D_{KL}(q_\phib(\z | \x) || p(\z))] + const.
\end{align}

\subsection{Our loss function}
We further set $\x_\psib(\z) = \x_\thetab(\z, 1)$ for parameter sharing. Then, our loss function is
\begin{align}
    &\min_{\thetab, \phib} \ I(q) + \beta D_{KL}(q_\phib(\z) || p(\z)) \\
    &\leq \mathbb E_{t, \x, \z \sim q_\phib(\x, \z)}\left[ \frac{1}{t^2} ||\x-\x_\thetab(\x_t(\x, \z), t)||_2^2 +\beta \frac{||\x_\thetab(\z, 1) - \x||_2^2}{2\sigma^2} + \beta D_{KL}(q_\phib(\z | \x) || p(\z))\right] + const \\
    &= \mathbb E_{t, \x, \z \sim q_\phib(\x, \z)}\left[ \bar \lambda(t) ||\x-\x_\thetab(\x_t(\x, \z), t)||_2^2 + \beta D_{KL}(q_\phib(\z | \x) || p(\z))\right] + const,
\end{align}
where $\bar \lambda(t)$ is $1/t^2$ if $t \neq 1$ and $\beta \delta(0)$ in $t = 1$ with Dirac delta function $\delta(\cdot)$. Empirically, we observe that setting $\bar \lambda(t)$ to $1/t^2$ for every $t$ leads to better performance.

\section{Implementation Details}
\label{appendix:implementation}
Table~\ref{appendix:tab:config} shows the training and architecture configuration we use in our experiments.
In our experiment, we directly parameterize the vector field $\mathbf v_\theta(\x_t, t)$ following \citet{liu2022flow}.
For MNIST and CIFAR-10 datasets, we employ DDPM++ architecture~\citep{song2020score} in the codebase of \citet{karras2022elucidating}\footnote{\href{https://github.com/NVlabs/edm}{https://github.com/NVlabs/edm}}. We evaluate FID using the code of \citep{karras2022elucidating}. We fix the random seed to $0$ throughout all experiments. We linearly increase the learning rate as in previous studies~\citep{karras2022elucidating,song2020score}. We use Adam optimizer with $\beta_1 = 0.9$, $\beta_2 = 0.999$, and $eps=1e-8$ for MNIST and CIFAR-10 datasets. Refer to our codebase for detailed configurations.

In \texttt{small} setting for encoder architecture, we use the MNIST generator architecture in Tab.~\ref{appendix:tab:config}, which is more than 20 times smaller than CIFAR-10 models.
For the distillation experiment, we use 500K pairs sampled from teacher ODEs. We find that student models overfit if the number of pairs is less than 500K.

For unconditional CIFAR-10 generation, we use two solvers -- RK45 and \heun. We set both \texttt{atol} and \texttt{rtol} to $1e-5$ for RK45 as in previous work~\citep{song2020score,liu2022flow}. We experiment with two configurations, config A and config B, and find that config A converges faster than config B at the expense of performance. Overall, our method converges faster than the independent coupling baseline.

\begin{table}[]
\centering
\caption{Architecture and training configurations. $^1$We use 200K and 300K iterations for $\beta=10$ and independent coupling, respectively. $^2$We use 500K and 600K iterations for $\beta=20$ and independent coupling, respectively.}
\vskip 0.15in
\begin{tabular}{@{}lccc@{}}
\toprule
                         & CIFAR-10 (A)         & CIFAR-10 (B)         & MNIST                \\ \midrule
Iterations               & varies$^1$              &   varies$^2$                   & 60K                  \\
Batch size               & 128                  & 128                  & 256                  \\
Learning rate            & $3e-4$               & $2e-4$               & $3e-4$               \\
LR warm-up steps         & 78125                & 5000                 & 8000                 \\
EMA decay rate           & 0.9999               & 0.9999               & 0.9999               \\
EMA start steps          & 300                  & 1                    & 300                  \\
Dropout probability      & 0.13                  & 0.13                 & 0.13                 \\
Channel multiplier       & 128                  & 128                  & 32                   \\
Channels per resolution  & $[2, 2, 2]$          & $[2, 2, 2]$          & $[2, 2, 2]$          \\
Xflip augmentation       & X                    & O                    & X                    \\
\# of params (generator) & 55.73M               & 55.73M               & 2.15M                \\
\# of params (encoder)   & 2.2M                 & 2.2M                 & 2.2M                 \\
\# of ResBlocks          & 4                    & 4                    & 2                    \\
$t$ range                & $[0,1]$              & $[1e-5, 1]$          & $[0, 1]$             \\
                         \bottomrule
\end{tabular}
\label{appendix:tab:config}
\end{table}

\section{Additional Results}
\label{appendix:additional}
\begin{figure}[]
\centering
\begin{tabular}{ccc}
     \raisebox{5em}{\rotatebox{90}{$\beta=10$}} \includegraphics[width = 0.3\textwidth]{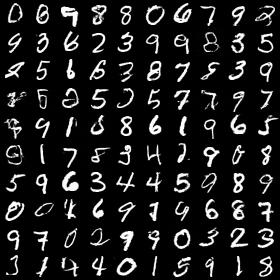} & \includegraphics[width = 0.3\textwidth]{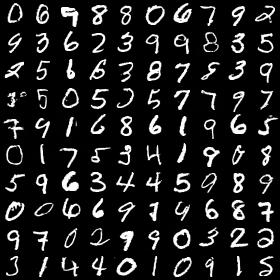} & \includegraphics[width = 0.3\textwidth]{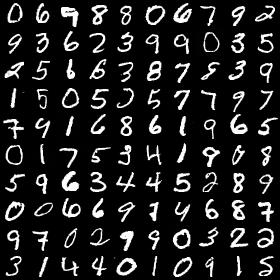}  \\

     \raisebox{5em}{\rotatebox{90}{$\beta=20$}} \includegraphics[width = 0.3\textwidth]{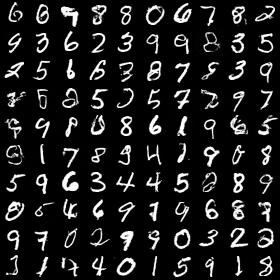} & \includegraphics[width = 0.3\textwidth]{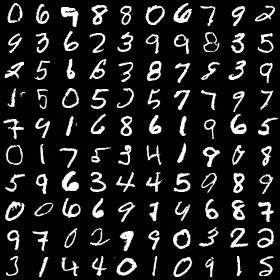} & \includegraphics[width = 0.3\textwidth]{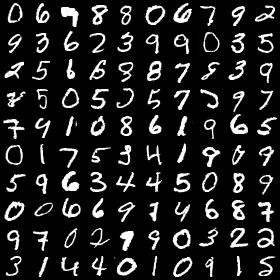}  \\

     \raisebox{5em}{\rotatebox{90}{independent}} \includegraphics[width = 0.3\textwidth]{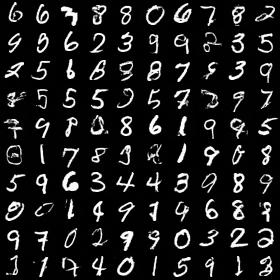} & \includegraphics[width = 0.3\textwidth]{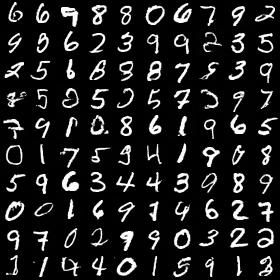} & \includegraphics[width = 0.3\textwidth]{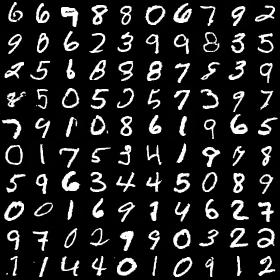}  \\

      NFEs = 5 & NFEs = 10 & NFEs = 128  \\
\end{tabular}
\caption{Uncurated MNIST samples.}
\label{appendix:fig:mnist}
\end{figure}

\begin{figure}
\centering
\begin{tabular}{ccc}
     \raisebox{5em}{\rotatebox{90}{$\beta=10$}} \includegraphics[width = 0.3\textwidth]{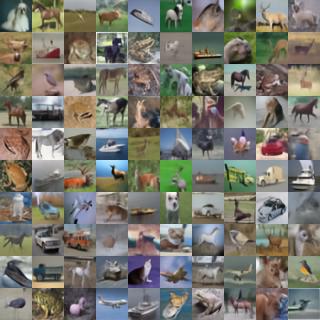} & \includegraphics[width = 0.3\textwidth]{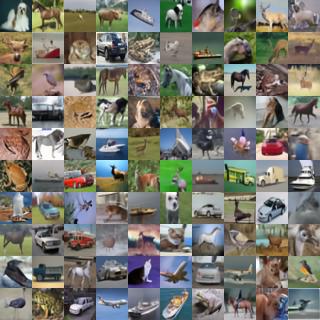} & \includegraphics[width = 0.3\textwidth]{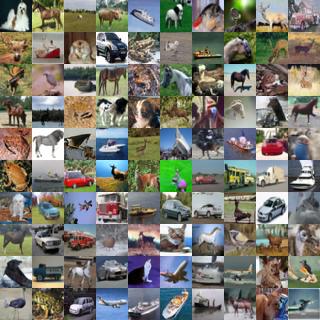}  \\

     \raisebox{5em}{\rotatebox{90}{$\beta=20$}} \includegraphics[width = 0.3\textwidth]{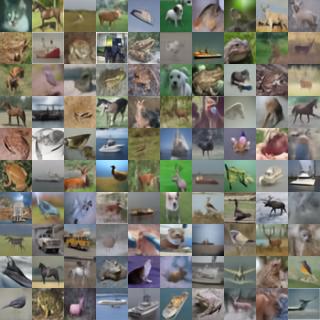} & \includegraphics[width = 0.3\textwidth]{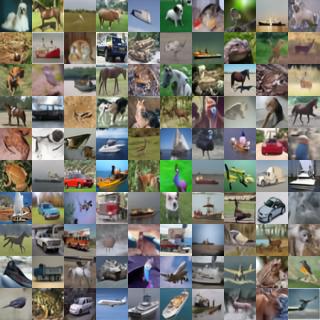} & \includegraphics[width = 0.3\textwidth]{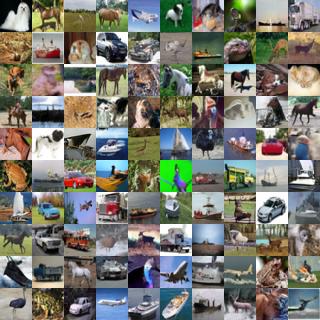}  \\

     \raisebox{5em}{\rotatebox{90}{independent}} \includegraphics[width = 0.3\textwidth]{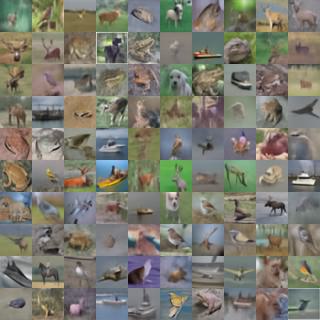} & \includegraphics[width = 0.3\textwidth]{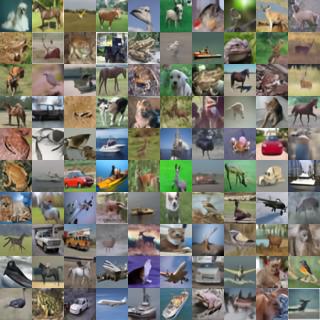} & \includegraphics[width = 0.3\textwidth]{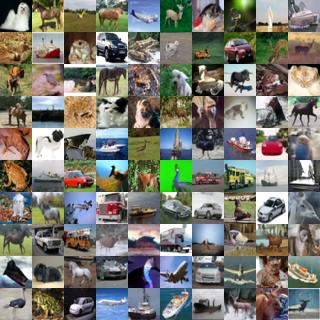}  \\

      NFEs = 5 & NFEs = 9 & full sampling  \\
\end{tabular}
\caption{Uncurated CIFAR-10 samples.}
\label{appendix:fig:cifar10}
\end{figure}

We further provide additional synthesis results of our method in Figs.~\ref{appendix:fig:mnist} and \ref{appendix:fig:cifar10}.

\end{document}